\documentclass{article}

\usepackage[preprint]{neurips_2024}

\usepackage[utf8]{inputenc} %
\usepackage[T1]{fontenc}    %
\usepackage{hyperref}       %
\usepackage{url}            %
\usepackage{booktabs}       %
\usepackage{amsfonts}       %
\usepackage{nicefrac}       %
\usepackage{microtype}      %
\usepackage{xcolor}         %
\usepackage{diagbox}

\usepackage{amsmath}
\usepackage{amssymb}
\usepackage{graphicx}
\usepackage{subfigure}
\usepackage{xcolor}
\usepackage{tabulary}
\usepackage{rotating} %
\usepackage{multirow} %
\usepackage{algorithm}
\usepackage{algorithmic}

\newcommand{\alg}{CANDERE-COACH~}
\newcommand{\fullalg}{\underline{\textbf{C}}lassifier \underline{\textbf{A}}ugmented \underline{\textbf{N}}oise \underline{\textbf{DE}}tecting and \underline{\textbf{RE}}labelling COACH~}

\title{CANDERE-COACH: Reinforcement Learning from Noisy Feedback}

\author{%
  Yuxuan Li \\
  Department of Computing Science \\
  University of Alberta \\
  Edmonton, Alberta, Canada \\
  \texttt{li.yuxuan@ualberta.ca} \\
  \And
  Srijita Das \\
  Computer and Information Science Department \\
  University of Michigan-Dearborn\\
  Dearborn, Michigan, USA \\
  \texttt{sridas@umich.edu} \\
  \AND
  Matthew E. Taylor \\
  Department of Computing Science \\
  University of Alberta \\
  Edmonton, Alberta, Canada \\
  \texttt{matthew.e.taylor@ualberta.ca} \\
}

\begin{document}

\maketitle

\begin{abstract}
In recent times, Reinforcement learning (RL) 
has been widely applied to many challenging tasks. 
However, in order to perform well, it requires access to a good reward function which is often sparse or manually engineered with scope for error.
Introducing human prior knowledge is often seen as a possible solution to the above-mentioned problem, such as imitation learning, learning from preference, and inverse reinforcement learning. Learning from feedback is another framework that enables an RL agent to learn from binary evaluative signals describing the teacher’s (positive or negative) evaluation of the agent's action. However, these methods often make the assumption that evaluative teacher feedback is perfect, which is a restrictive assumption. In practice, such feedback can be noisy due to limited teacher expertise or other exacerbating factors like cognitive load, availability, distraction, etc. In this work, we propose the CANDERE-COACH algorithm, which is capable of learning from noisy feedback by a nonoptimal 
teacher. We propose a noise-filtering mechanism to de-noise online feedback data, thereby enabling the RL agent to successfully learn with up to 40\% of the teacher feedback being incorrect.
Experiments on three common domains demonstrate the effectiveness of the proposed approach.
\end{abstract}

\section{Introduction}

Reinforcement learning (RL) has made rapid progress in part due to the advancements of deep learning, which has been successfully applied to the challenging game of Go~\cite{silver2017mastering}, solving Rubik's cube with a robot arm~\cite{akkaya2019solving}, and for deciding the treatment regimen for cancer~\cite{tseng2017deep}. Although successful in solving these problems, its progress has still been significantly hindered by the well-known sample-inefficiency problem~\cite{ibarz2021train} because the agent may need millions of interactions with an environment to learn a near-optimal policy.

Moreover, deep RL's performance often deteriorates in domains with sparse reward because of slower propagation of the reward signal to the entire state-space~\cite{yu2020meta, knox2023reward}. To combat this problem, reward functions may be manually specified by task experts or RL developers, often by trial-and-error. 
However, even when human subjects carefully designed reward functions, errors can lead to reward hacking~\cite{laidlaw2024preventing} or other unintended behaviors. 
To address sample inefficiency and problems related to reward design, human-in-the-loop RL~\cite{retzlaff2024human} has been used to guide RL algorithms. 
Human prior knowledge has been used in the form of demonstrations~\cite{schaal1999imitation}, action-advising~\cite{torrey2013teaching},  policy-shaping~\cite{griffith2013policy}, and action advising~\cite{torrey2013teaching}, to name a few. To address the reward design problem, human advice has been used to learn the reward model in various frameworks, such as inverse reinforcement learning~\cite{ng2000algorithms}, learning from preference~\cite{christiano2017deep,lee2021pebble}, learning from human feedback~\cite{knox2009interactively, macglashan2017interactive}, etc. Deep COACH~\cite{arumugam2019deep} is one such learning from feedback paradigm in which the teacher observes an agent in action and provides scalar feedback denoting agreement or disagreement. This feedback is in turn used as an advantage function in the RL algorithm. While this method has been successful in training RL agents without a reward function, one major assumption is that the feedback is perfect (e.g., noise-free). However, collecting feedback from humans is an expensive process which demands attention, time, and focus from the human. Hence, this assumption is restrictive and the collected feedback might be noisy. Addressing this critical shortcoming and making feedback learning more useful in real-world settings is the primary motivation of this work.

While there has been much work in the RL literature to handle imperfect human knowledge (particularly with respect to demonstrations~\cite{chen2021learning}), learning from noisy binary feedback is an important unmet need.
Contrary to this, identifying noisy  data~\cite{han2018co, younesian2021qactor} and outlier detection~\cite{liu2013svdd} has been widely studied in supervised learning.
Motivated by these advances, we propose a policy learning from feedback framework where 1) the agent doesn't have access to the reward function 2) can learn from \textit{noisy feedback} given by the human/agent teacher as positive and negative feedback 3) the agent learns a policy to maximize the likelihood of doing what the teacher wants. Our proposed framework consists of a \textit{classifier augmented noise detecting module} that is capable of detecting and filtering noisy feedback, which is further used to guide the RL agent to solve the task. 
\\
\noindent \textbf{Contributions} of this paper include:
(1) a demonstration that well-known learning from feedback methods like Deep-COACH can fail to learn from noisy, limited teacher feedback 
(2) a novel algorithm that includes a learning model to identify and filter noise, 
(3) investigating the behavior of the proposed algorithm with varying feedback noise levels, and 
(4) an empirical evaluation on three domains showing that our proposed method learns from teacher feedback with up to 40\% noise, significantly outperforming the relevant baselines. (5) potential for using our method like a plug and play tool with other learning from feedback methods like Deep TAMER.

\section{Related Work}
\textbf{RL agents learning from teacher advice:}
RL agents are
often challenged by sparse reward domains. 
Receiving help from a knowledgeable teacher can ameliorate such problems, such as when a teacher agent (or human) provides demonstrations (e.g., in imitation learning ~\cite{billard2003discovering,giusti2015machine,ross2011reduction} or inverse reinforcement learning~\cite{das2020model,ho2016generative}). 
There are also teacher-student frameworks~\cite{torrey2013teaching,ilhan2021action} where the student agent asks for action advice from the teacher, and learning from preferences~\cite{wilson2012bayesian,christiano2017deep,lee2021pebble}, where preferences over pairs of trajectories are provided.
Learning from feedback is another advising framework, where agents learn from binary feedback provided by the teacher.
Knox et al.~\cite{knox2009interactively} proposed TAMER to exploit human feedback signals; which in turn is used to learn an expectation of human feedback that the agent can maximize. %
Macglashan et al.~\cite{macglashan2017interactive} proposed COACH, which uses feedback as the advantage function in the policy-gradient objective. These methods have also been extended in Deep RL settings like Deep TAMER ~\cite{warnell2018deep} and Deep COACH~\cite{arumugam2019deep}.
Loftin et al.~\cite{loftin2016learning} proposed methods that take teacher's strategy into account by inferring teacher's strategy, but most of the experiments are done in bandit domains.
Compared to learning from demonstration methods, learning from feedback may require less knowledge, as judging good or bad actions is essentially easier than specifying the optimal action.
However, to the best of our knowledge, all previous work on learning from feedback does not explicitly try to detect and correct errors. We propose a learning from feedback method that learns from \textit{noisy teacher feedback}.\\ 
\textbf{Noise detection in supervised learning:}
Learning with noisy data has become increasingly challenging with massive datasets and increasing the risk of noise from annotated data collection~\cite{song2022learning}. 
In supervised learning, noise can be easily memorized by the deep networks \cite{zhang2021understanding}, hurting generalization performance. 
Several learning methods have been proposed to detect noisy labels.
MentorNet \cite{jiang2018mentornet} uses a separate mentor network to guide the classifier with a curriculum of noisy labels.
For semi-supervised anomaly detection, Chen et al. \cite{chen2015robust} use SVDD as a one-class classifier to detect outliers.
Han et al.~\cite{han2018co} proposed Co-teaching, an example of an unsupervised anomaly detection method. 
Co-teaching maintains two deep networks and lets them select data based on their respective losses and feed outputs into each other reciprocally, allowing learning from noisy data. 
Similar ideas of selecting data based on losses can be found in QActor~\cite{younesian2021qactor}, where detected noisy labels are further corrected by oracle during active learning.
Motivated by progress in this direction, our proposed work adapts loss-based noise detection techniques for deep networks into learning from feedback framework for training deep RL agents from noisy feedback.

\section{Background}
\textbf{Reinforcement Learning:} RL is defined using a Markov Decision Process (MDP).
An MDP is denoted by a quintuple as $M=\{\mathcal{S}, \mathcal{A}, \mathcal{T}, r, \gamma \}$, where $\mathcal{S}$ denotes the agent's state space,  $\mathcal{A}$ is the agent's action space, $\mathcal{T}:\mathcal{S}\times \mathcal{A} \times \mathcal{S} \rightarrow [0,1]$ is the environmental dynamics transition probability, 
$r: \mathcal{S} \times \mathcal{A} \times \mathcal{S} \rightarrow \mathbb{R}$ is the function that gives an immediate reward, $\gamma$ is a discount factor.
In a typical MDP setting, the RL agent starts at state $s_0$ and takes an action $a_0$ following its policy $\pi$, which leads the agent to its next state $s_1$ and receives reward $r_0$. The interaction repeats for $T$ steps until a terminal state is reached.  
RL agent optimizes the policy $\pi$ by maximizing the expected cumulative reward $\mathbf{E}_{\tau \sim \pi} [ \Sigma_{t=0}^{T} r_t]$ 

\noindent \textbf{Deep COACH:}
Feedback is often defined as a scalar value $f$, describing the teacher's judgment of the agent's current behavior, and the teacher can be a human teacher or a scripted teacher,
Like COACH~\cite{warnell2018deep} and TAMER~\cite{knox2009interactively}, 
we define feedback as $f \in \{-1, 1\}$, where $-1$ means the teacher discourages the agent's behavior, and $1$ encourages the agent's behavior, which is a pair of state and action $\langle  s_t, a_t\rangle  $, at time step $t$.
In COACH, the feedback $f_t$ is deemed as an estimate of the advantage value and is used to update the policy.
In Deep-COACH~\cite{arumugam2019deep},  the policy is updated as Equation~\ref{equ:coach}:
\begin{equation}
    \label{equ:coach}
    \nabla_{\theta_t} {J}(\theta_t)= \mathbf{E}_{a \sim \pi^h_{\theta_t}(\cdot|s_t)}[\nabla_{\theta_t} log (\pi_{\theta_t}(a_t|s_t))\cdot f_t].
\end{equation}

\section{Classifier Augmented Noise Detecting and Relabelling COACH}

\subsection{Problem Statement}
\noindent {\bf Given:} An RL agent that has no access to the reward function $r$, a teacher $T$ that can provide noisy binary feedback $\hat{f} \in \{-1, 1\}$ based on the learning agent's visited state and action; and the ground truth feedback $f \in \{-1, 1\}$.\\
\noindent {\bf Objective:} Train the  agent policy $\pi_{\theta}$ by using the noisy feedback $\hat{f}$ given by the teacher.\\
\noindent \textbf{Assumptions:} We make the following assumptions: (1) Static and symmetric noise is added to $f$ to produce $\hat{f}$ (i.e., the noise distribution 
does not change over time and every feedback can be incorrectly flipped with equal probability.  (2) The proportion of labels that are flipped (i.e., the noise proportion $p_{noise}$) is known.\footnote{We relax this assumption later in Effect of noisy pretraining Dataset, Section 5.2.} (3) The ground truth feedback is deterministic for each state and action pair, i.e., only one correct feedback exists for each state-action pair. (4) The agent does not have access to the reward function during training (but may be used to evaluate the agent's policy).

\subsection{Methodology and Algorithm}
\begin{figure}[h]
    \centering
    \includegraphics[keepaspectratio=true, width=0.5\textwidth]{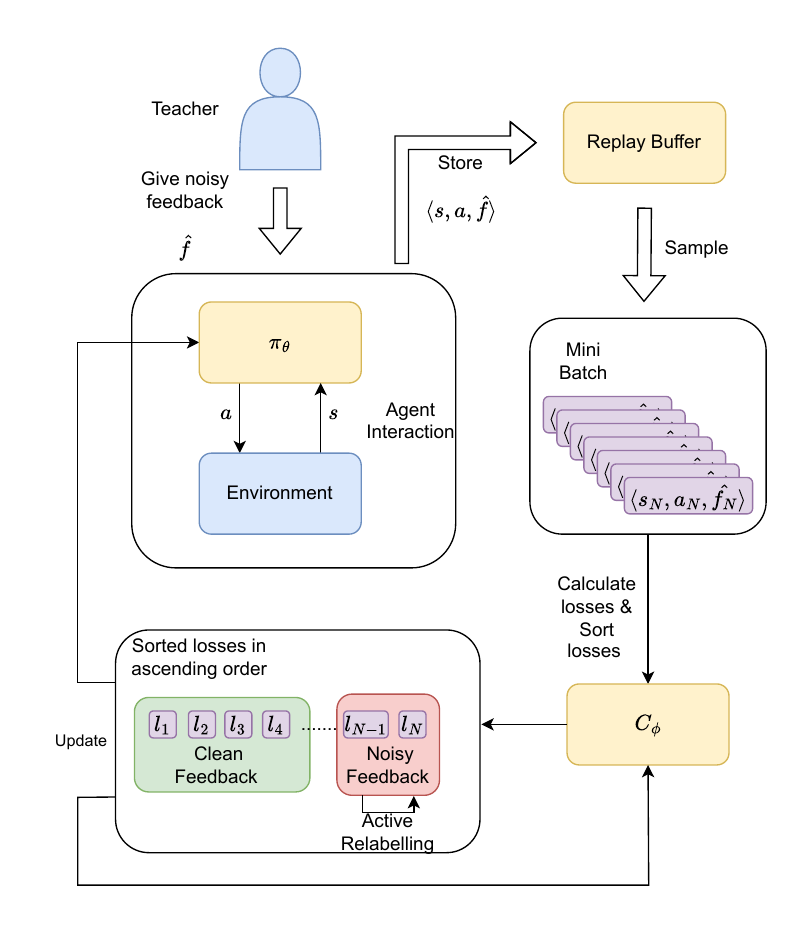}
    \caption[CANDERE-COACH overview] {The overview of \alg. We use a classifier $C_\phi$ to filter noisy feedback and update policy $\pi_\theta$ and $C_\phi$ with filtered minibatches. }
    \label{fig:NR-HITLRL}
\end{figure}

The proposed framework \fullalg (CANDERE-COACH) is illustrated in Figure~\ref{fig:NR-HITLRL}. The agent follows the policy $\pi_{\theta}$. The teacher provides noisy binary feedback $\hat{f}$ on the state-action pairs visited by the agent. This feedback, along with the corresponding state-action pair, is stored in a replay buffer. During training, a mini-batch is sampled and is input to a classifier, $C_{\phi}$ (Section 4.2.1), which acts as a noise detector. This classifier is trained (as the agent learns) and identifies correct and incorrect feedback based on its loss function (Section 4.2.2). We also employ a method called `Active Relabeling' (Section 4.2.3) to convert some of the feedback suspected to be incorrect.
The aggregation of the original and relabelled feedback constitutes the filtered batch --- this batch becomes input for the agent's policy training and the training of the classifier.\\
CANDERE-COACH comprises the below components:\\

\noindent \textbf{4.2.1 \underline{Noise-filtering classifier}}:
In order to identify noisy feedback, we train a classifier $C_\phi: S\times A \rightarrow [0,1]$,
\footnote{The output is actually a vector of predicted probabilities for positive and negative feedback that sums to 1.}
which maps the state and action pair to predicted feedback probability distribution. The feedback dataset $\mathcal{D}=\{(s_i,a_i,\hat{f_i})\}_{i=1}^{N}$ where $(s_i$, $a_i)$ refers to the $i^{th}$ state, action pair and $\hat{f_i}$ refers to the corresponding observed feedback.
 This classifier is pretrained on a small noise-free feedback dataset $\mathcal{D}$ using the cross entropy loss.
as shown in Eqn~\ref{equ:cel}), where $N$ refers to the size of the pretraining dataset, $P(c=\hat{f_i}|s_i,a_i,\phi)$ refers to the predicted probability of feedback by the classifier and $c \in \{-1, +1\}$ is the feedback class.

\begin{equation}
    l(\phi)  = -\sum_{i=1}^{N} \sum_{c \in \{-1, 1\}} \mathbb{I}[\hat{f_i}=c]\log {P(\hat{f_i}=c|s_i,a_i,\phi})
\label{equ:cel}
\end{equation}

We also use focal loss \cite{lin2017focal} instead of cross entropy loss for pretraining datasets of imbalanced classes, as shown in Eqn~\ref{equ:focallossalpha}. $\gamma$ is the focusing parameter and $\alpha: f \rightarrow \mathbb{R}$ is a map from a data class of input to a scalar weight, which is set based on the proportion of that class in the dataset. 
\label{equ:focallossalpha}
\begin{multline}
    l(\phi)  = -\sum_{i=1}^{N} \sum_{c \in \{-1, 1\}} \mathbb{I}[\hat{f_i}=c]\log {P(\hat{f_i}=c|s_i,a_i,\phi}) \alpha(c)(1-P(\hat{f_i}=c|s_i,a_i,\phi))^{\gamma}
\end{multline}

The classifier $C_{\phi}$ is retrained after every step of noise detection and active relabeling, described in detail in the next section.

\noindent \textbf{4.2.2\underline{ Noise-detection}}: In this step, we detect potentially noisy feedback by using the trained classifier $C_{\phi}$. In every learning iteration, once a mini-batch is sampled from the replay buffer, $C_{\phi}$ predicts the point-wise loss 
of each state-action-feedback tuple which are then sorted in increasing order as per their loss values.  We treat data points with small losses as clean data. This is because neural networks can learn clean data in the early stages of training ~\cite{han2018co} because estimating the distributions of these data-points is relatively easier. Mathematically, clean data batch $B_c$ as identified by the classifier is filtered as mentioned below where $l_{B'}(\phi)$ is the cross-entropy loss of the classifier, and $R(B)$ refers to the remember rate that specifies how many data points needs to be included in the batch.
\begin{equation}
    \label{equ:sort}
    B_c=\operatorname*{arg\,min}_{B':|B'|>= R(B)|B|}l_{B'}(\phi)
\end{equation}
In the above equation, we set remember rate $R(B) = 1 - p_{noise}$.
After $B_c$ is identified, we sort the data points with larger loss values and consider them as potentially noisy data because deep networks cannot estimate noisy distributions in the early stages of training owing to these conditional distributions being harder to estimate~\cite{han2018co}. Hence, the suspected noisy batch $B_n$ as identified by the classifier is estimated as below:
\begin{equation}
    \label{equ:active_relabelling}
    B_n=\operatorname*{arg\,max}_{B':|B'|>= R'(B)|B|}l_{B'}(\phi)
\end{equation}
In the above equation, 
$R'(B)$ refers to a hyperparameter that decides how many data points needs to be included in $B_n$. 
\\
\noindent \textbf{4.2.3 \underline{Active relabeling}}: Once suspected noisy batch $B_n$ has been identified, we flip the feedback labels of the batch $B_n$ resulting in $B_{ar}$. The intuition is that the opposite feedback label is necessarily the correct label for binary valued feedback. We call this method \textit{active-relabelling}. 
After data has been relabelled, $B_c$ and $B_{ar}$ are provided to the RL agent for training the policy $\pi_{\theta}$ as well as to the classifier $C_\phi$ for online training as per Eqn~\ref{equ:cel}. We use policy gradient to train the RL agent using the filtered feedback as per Eqn~\ref{equ:coach}.\\

\noindent \textbf {Algorithm:} Algorithm~\ref{alg:activerelabelling} shows the algorithm for CANDERE-COACH. 
With a pretrained classifier $C_\phi$, the agent interacts with the environment and queries the teacher for feedback at a set frequency (line 5). 
The agent's observation, action, and feedback are stored as a tuple in the replay buffer $R$.
After sampling a minibatch $B$ from R,  it is fed into the classifier to evaluate data point-wise cross entropy loss (line 8).
These loss values are sorted in ascending order and the first 
$R(B)|B|$
data-points in the batch are selected as $B_{c}$ (line 10). 
Similarly, $R'(B)|B|$ data-points (denoted as $B_{n}$) are selected by sorting them per their cross entropy loss value (Equation~\ref{equ:cel}) in descending order and the feedback label of this set is flipped (lines 11--12) as $B_{ar}$. 
In the last step, lines 14--15, in addition to updating the policy $\pi_\theta$, the classifier is also updated using online training with the filtered feedback batch ($B_{c}$ and $B_{ar}$).
The classifier gradually adapts to the new state distribution as in the training set and related ablation study can be found in Section~\ref{sec:abaltion_studies}.

\begin{algorithm}[h]
  \caption{\fullalg}
  \label{alg:activerelabelling}
  \textbf{Input}: {Pretrained Classifier $C_\phi$, Policy $\pi_\theta$, Teacher $T$, Noise Proportion $p_{noise}$, Maximum Episode Length $l$, Maximum \# of episode $N_e$, Replay Buffer $R$ size $N$, Batch size $b$, Label flipping rate $R'(B)$, Remember Rate $R(B)$ 
 
}
    \begin{algorithmic}[1]
    \STATE Initialise random policy $\pi_\theta$
    \STATE Initialise empty replay buffer $R$

    \FOR{$i \leftarrow 1,2,...,N_e$}
    \FOR{$j \leftarrow 1,2,...,l$}
    \STATE Agent with policy $\pi_\theta$ interacts with the environment and queries teacher $T$ for noisy feedback $\hat{f}$
    \STATE Sample batch $B=\{\langle s_0,a_0,f_0 \rangle, ... \langle s_{b-1},a_{b-1},f_{b-1} \rangle\}$ from $R$
    \STATE Use Classifier $C_\phi$ to predict $f$ on state-action pairs in $B$ 
    \STATE Calculate the cross entropy loss $L=\{l_0, ... l_{b-1}\}$ for each data sample in $B$ following Equation~\ref{equ:cel}
    \STATE Sort $B$ in ascending order based on $L$
    \STATE Pick $R(B)|B|$ items with minimised losses from $B$ as $B_{c}$, following Equation~\ref{equ:sort} 
    \STATE Pick $R'(B)|B|$ items with maximised losses as $B_{n}$, following Equation~\ref{equ:active_relabelling}
    \STATE Flip feedback labels of $B_{n}$ as $B_{ar}$
    \STATE Train $\pi_\theta$ with $B_{c}$ and $B_{ar}$ following Equation~\ref{equ:coach}
    \STATE Train Classifier $C_\phi$ with $B_{c}$ and $B_{ar}$ following Equation~\ref{equ:cel} 
    
    \ENDFOR
    \\
    \ENDFOR
    \STATE return $\pi_\theta$ 
    \end{algorithmic}
\end{algorithm}

\section{Experimental evaluation and results }

\noindent \textbf{Research questions} This section addresses the following three questions:
\begin{enumerate}
\item{} In what kind of setup is the performance of Deep COACH sensitive to noisy feedback?
\item {} Can CANDERE-COACH learn effectively with different proportions of noisy feedback? 
\item {} Does active relabelling help CANDERE-COACH in noise correction?

\end{enumerate}

\begin{figure}[h]
    \centering
  \subfigure[]
  {\label{fig:cart_pole}
  \includegraphics[trim={5cm 0 5cm 3cm},clip, keepaspectratio=true, height=2.5cm]{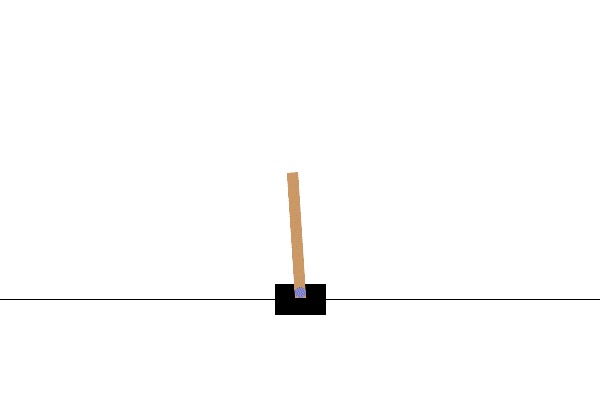}}
  \subfigure[]
  {\label{fig:door_key}
  \includegraphics[keepaspectratio=true, height=2.5cm]{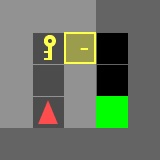}}
  \subfigure[]
  {\label{fig:lunar_lander}
  \includegraphics[trim={3cm 1cm 3cm 0},clip,keepaspectratio=true, height=2.5cm]{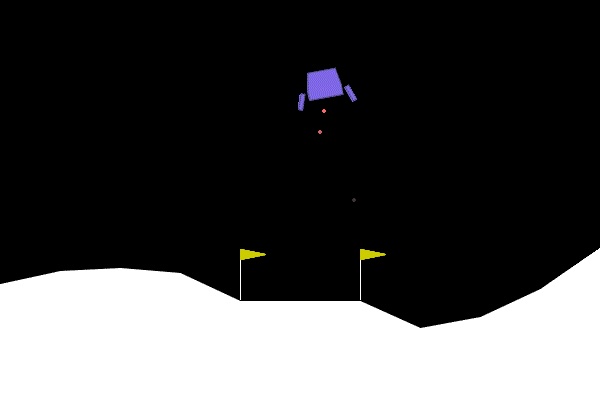}}

    \caption[Our domains] {(a) Cart Pole,  (a) Minigrid Door Key, and (c) Lunar Lander are used for evaluation} 
    \label{fig:domains}
\end{figure}

\noindent{\bf Domains:}
We conduct our experiments in three Gymnasium \cite{towers_gymnasium_2023} domains: Cart Pole, Lunar Lander, and Minigrid Doorkey, 
as shown in Figure~\ref{fig:domains}. 
In Cart Pole, the agent controls a moving cart to prevent the attached rod from falling. 
The Lunar Lander agent has to land a spacecraft on the moon, controlling three thrusters to avoid crashing.
In Minigrid Door Key, the agent needs to explore to find a key, unlock the door, and reach the goal.

\noindent{\bf Evaluation metrics:}
The agent is evaluated based on its accumulated reward on evaluation episodes by executing the current policy. 
During evaluation episodes, the agent's policy is frozen, hence no exploration.
Recall that the agent does not have access to the reward signal --- however, %
we use the default built-in reward function of our domains to evaluate the agent performance. 
We also use the $\%$ of correct feedback of the filtered data, namely the \textit{pure ratio} to evaluate the algorithm's noise-filtering capability.

\noindent{\bf Experimental Settings:} The feedback is provided by a scripted teacher.
The ultimate goal of this algorithm is to learn from noisy human feedback, a fixed teacher allows for better evaluation rigor and repeatability. The teacher uses a pre-trained 
expert policy for each domain, providing negative feedback when the agent fails to choose the optimal action, and positive feedback otherwise. 
In our settings, Just like real human teachers who cannot provide feedback at every step, the agent receives feedback at fixed intervals of time steps. \footnote{The details of feedback frequency can be found in Appendix Section~\ref{sec:hyperparam}.} 
The feedback's noise is symmetric, i.e., all feedback labels are randomly and independently flipped by a fixed probability depending on the noise ratio which is less than 50\%. 
The maximum number of feedback that an agent can receive is defined as its budget. 
CANDERE-COACH has a pretrained classifier that uses a small non-noisy feedback dataset and also has access to a limited number of noisy feedback (budget). \\
\noindent \textbf{Baselines:}The two baselines that we compare CANDERE COACH against are: (1) Deep COACH, which is allowed the same amount of noise-free feedback budget at the beginning of an episode as CANDERE COACH; this budget is the same as the pretraining dataset of our algorithm but consists of random states and actions that the agent visited
(2) Deep COACH (Preload), which loads the exact same dataset for classifier pretraining into the replay buffer as CANDERE-COACH. 
Noticeably, PEBBLE~\cite{lee2021pebble} does not serve as a baseline since it requires preferences over trajectories, while in our settings, the teacher provides feedback towards a set of state and action.
In our experiments, the baselines always have the same number of feedback budget to ensure a fair comparison.
Each experiment consists of 10 runs of different seeds. 
\footnote{More experimental settings and hyperparameter details can be found in Appendix Sections~\ref{sec:hyperparam}.}

\subsection{Results}
\label{sec:noise_evaluation}
\begin{figure*}[h]
  \centering
  \subfigure[Unlimited budget of feedback ]
  {
  \label{fig:coachdifferentnoise}
  \includegraphics[width=0.45\textwidth]{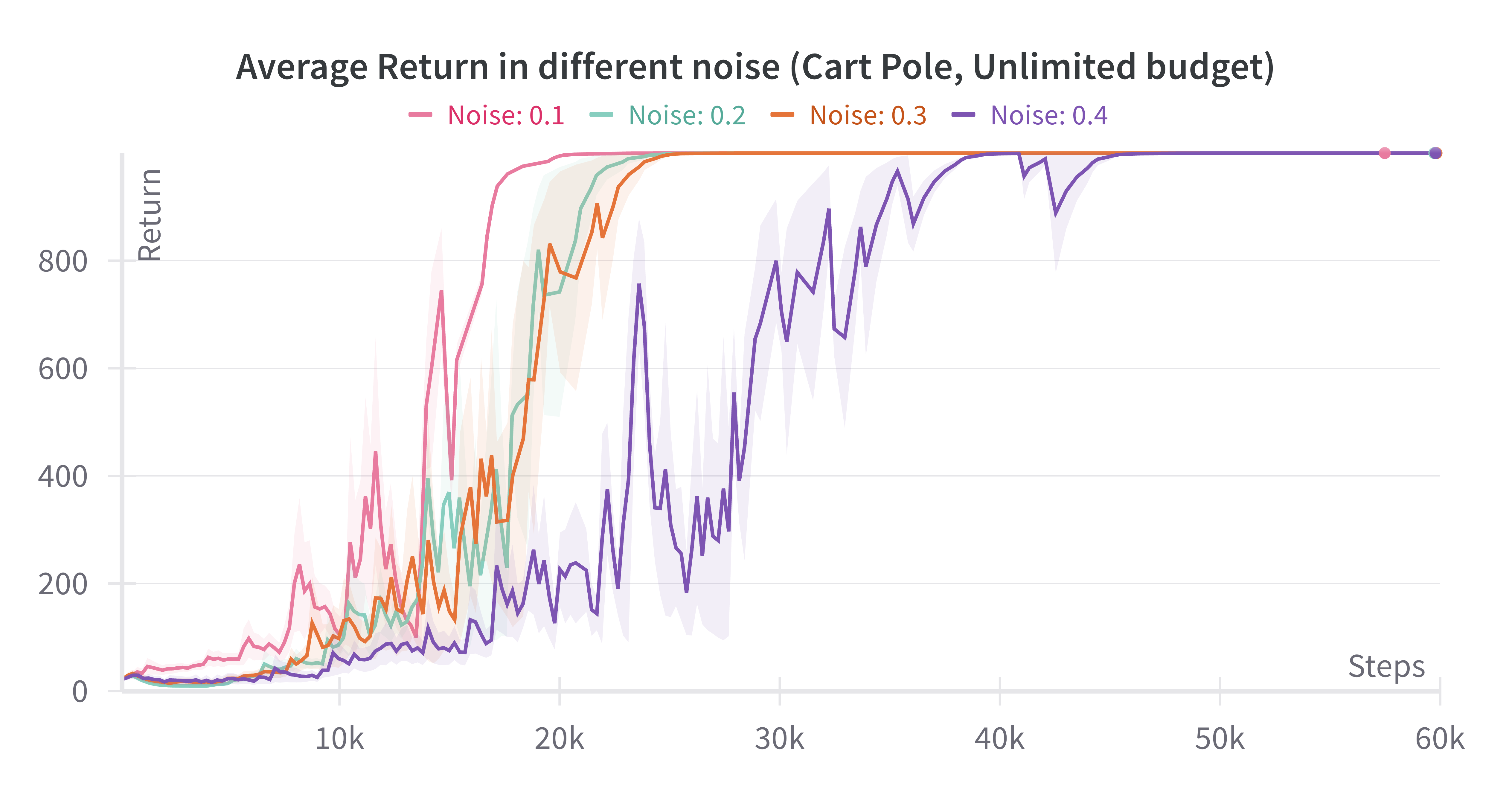}
  }
  \subfigure[Limited budget (1000) of feedback]{\label{fig:coachdifferentnoiselimited_budget}
  \includegraphics[width=0.45\textwidth]{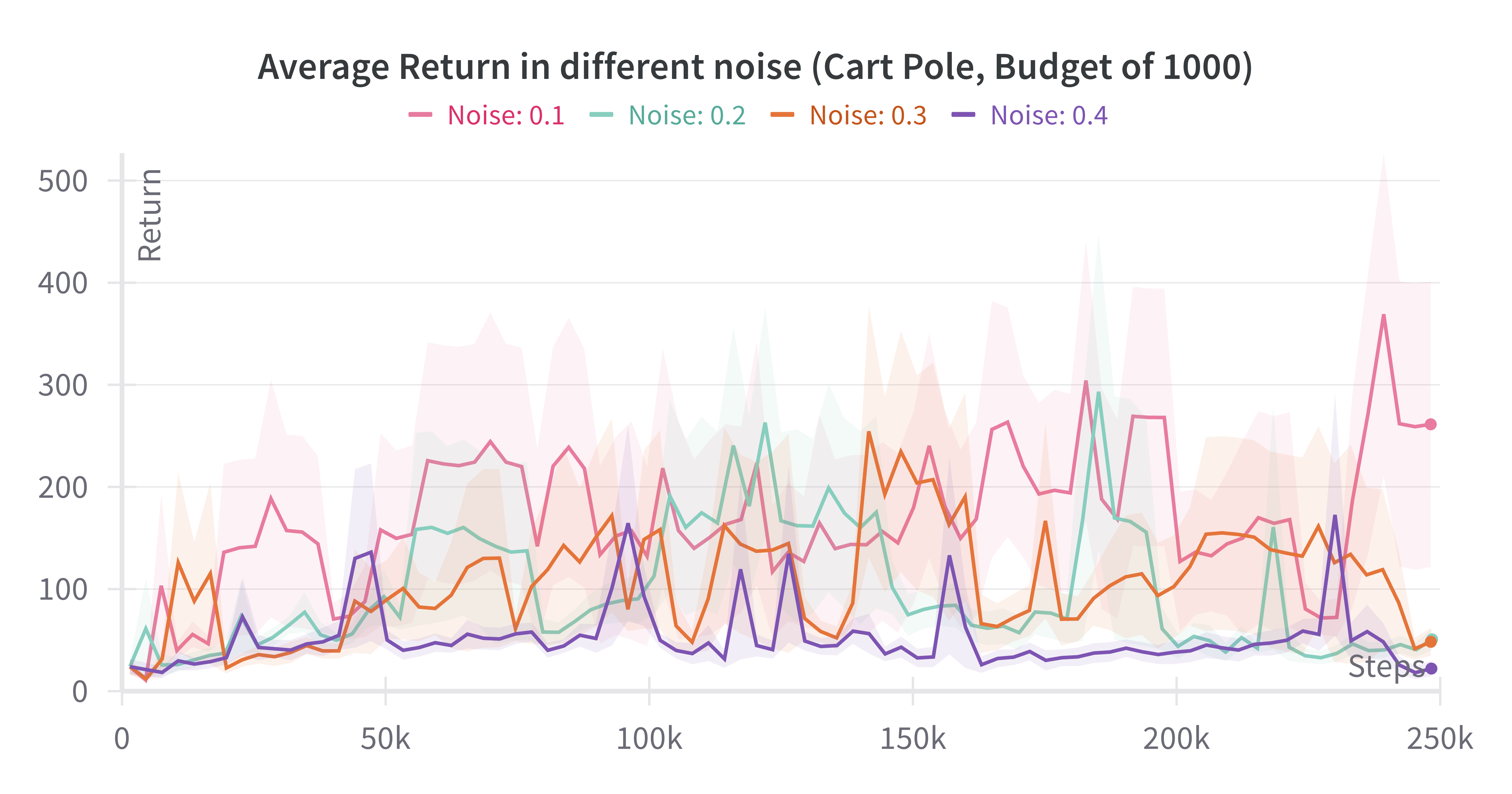}}
\caption{Performance of Deep COACH under different scales of noises in Cart Pole. While with an unlimited budget the Deep COACH is even able to learn against 40\% noise slowly, the performance of Deep COACH significantly deteriorates with a limited budget.}
\label{fig:exp1}
\end{figure*}

\noindent \textbf{When is Deep COACH sensitive to noisy feedback? }
To answer \textbf{RQ1}, we conduct experiments to evaluate Deep COACH with different proportions of noise.
We evaluated Deep COACH with noise amounts of \{0\%, 10\%, 20\%, 30\%, 40\%\} in the three domains. We consider both limited and unlimited feedback settings for this experiment.
As shown in Figure~\ref{fig:coachdifferentnoise} and Figure~\ref{fig:coachdifferentnoiselimited_budget} for Cart Pole, 
we observe that higher noise leads to worse agent performance (less reward) and/or more time required to converge to optimal performance.
Interestingly, with an unlimited budget, the agent is still able to learn with 40\% feedback noise.
Statistically, by the law of large numbers, as long as there are more correct labels than incorrect ones and unlimited feedback data, the agent eventually learns the correct policy given infinite amounts of feedback and learning time. 
However, for limited budget feedback,
as shown in Figure~\ref{fig:coachdifferentnoiselimited_budget}, the agent performance significantly deteriorates and even unlearns over time. 
Similar results can also be found in the other two domains (Lunar Lander and Minigrid Doorkey), shown in Appendix Section~\ref{sec:deep_coach_with_noise}. 

To summarize, we address \textbf{RQ1} by showing that noisy feedback poses a significant negative impact on Deep COACH, especially in a \textit{limited feedback setting}.

\noindent \textbf{CANDERE-COACH evaluation:}
\label{sec:limited_budget_alg}
To answer \textbf{RQ2} and \textbf{RQ3}, we evaluate the performance of the original \alg with a limited feedback budget based on the performance of Deep COACH from the previous experiments.
We also evaluate CANDERE-COACH without active relabelling, denoted as CANDERE-COACH (w/o AR).

\begin{figure*}[h]
    \centering
      \subfigure[]
      {\label{fig:all_plot_cp_n30}
      \includegraphics[width=0.32\textwidth]{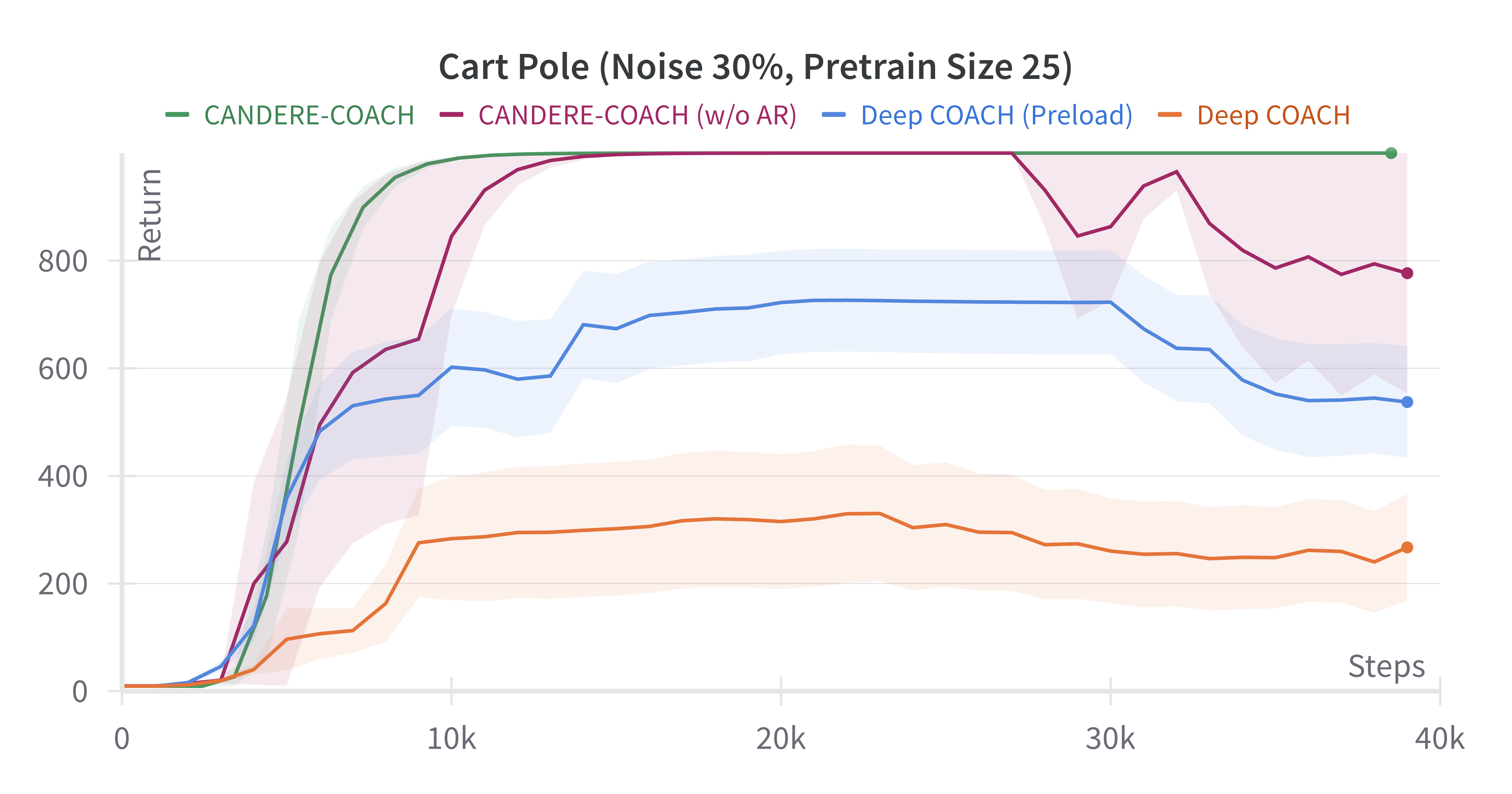}}
     \subfigure[]{\label{fig:all_plot_dk_n30}
      \includegraphics[width=0.32\textwidth]{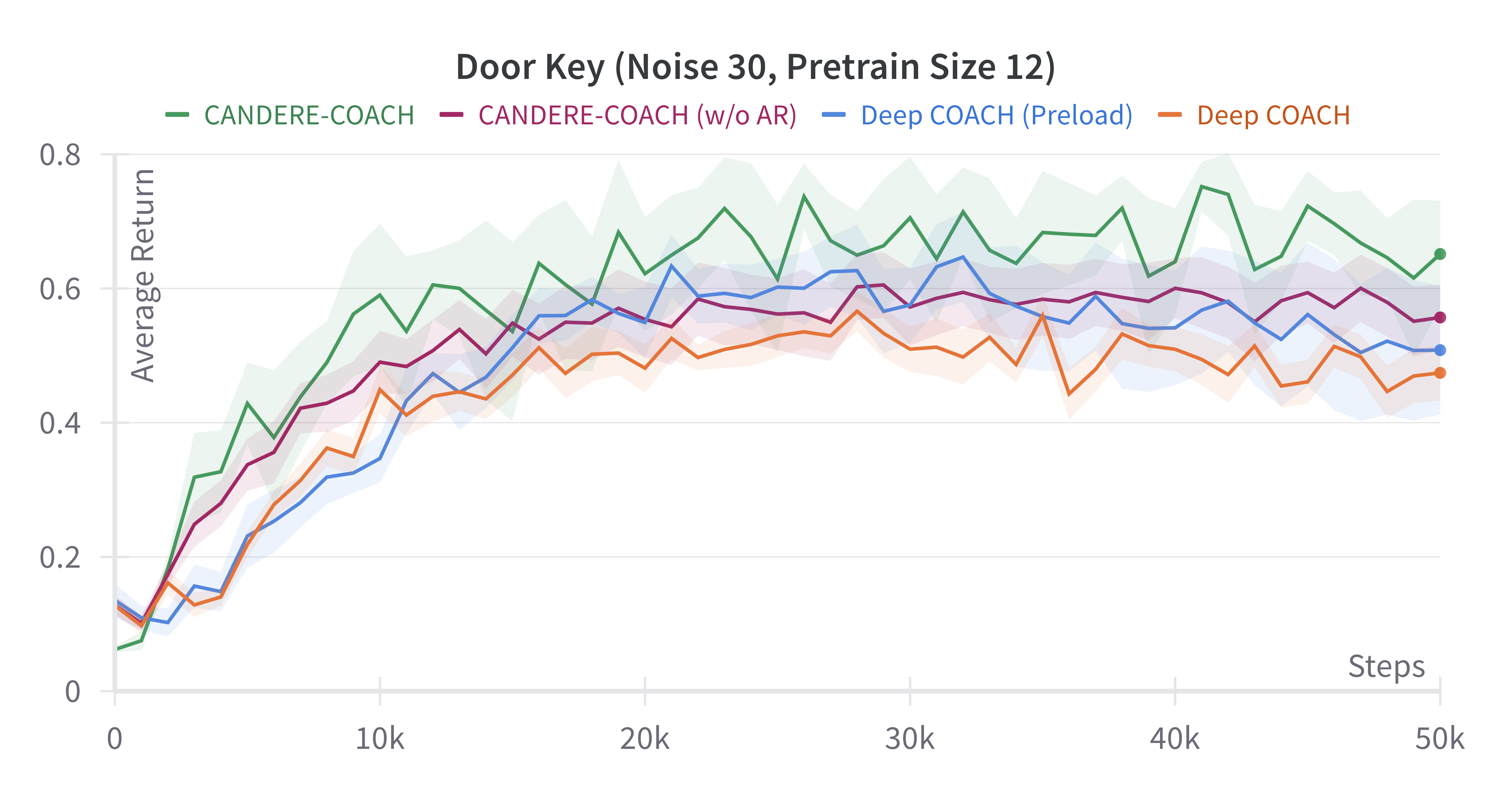}}
    \subfigure[]{\label{fig:all_plot_ll_n30}
      \includegraphics[width=0.32\textwidth]{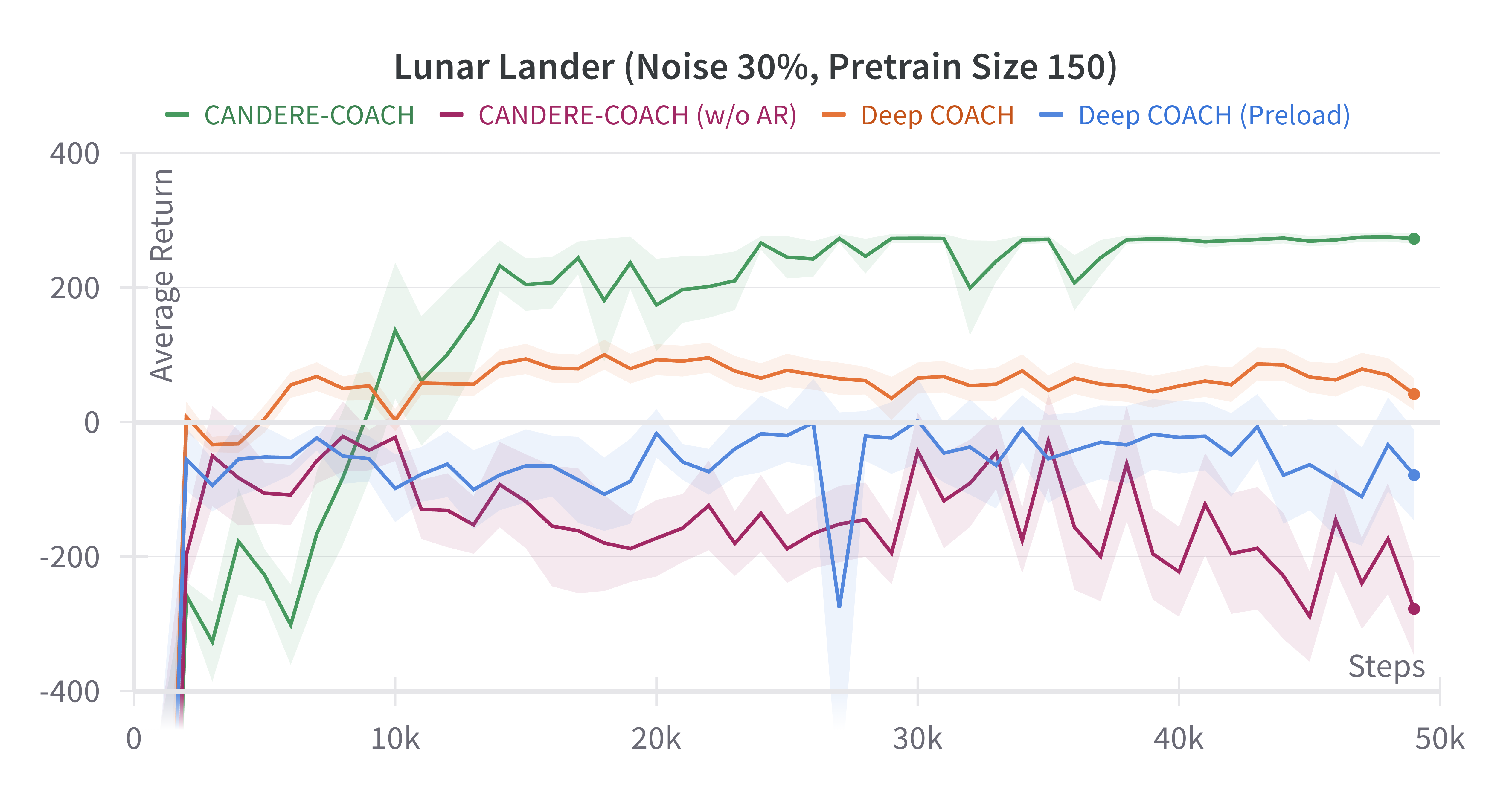}}
    \subfigure[]
      {\label{fig:all_plot_cp_n40}
      \includegraphics[width=0.32\textwidth]{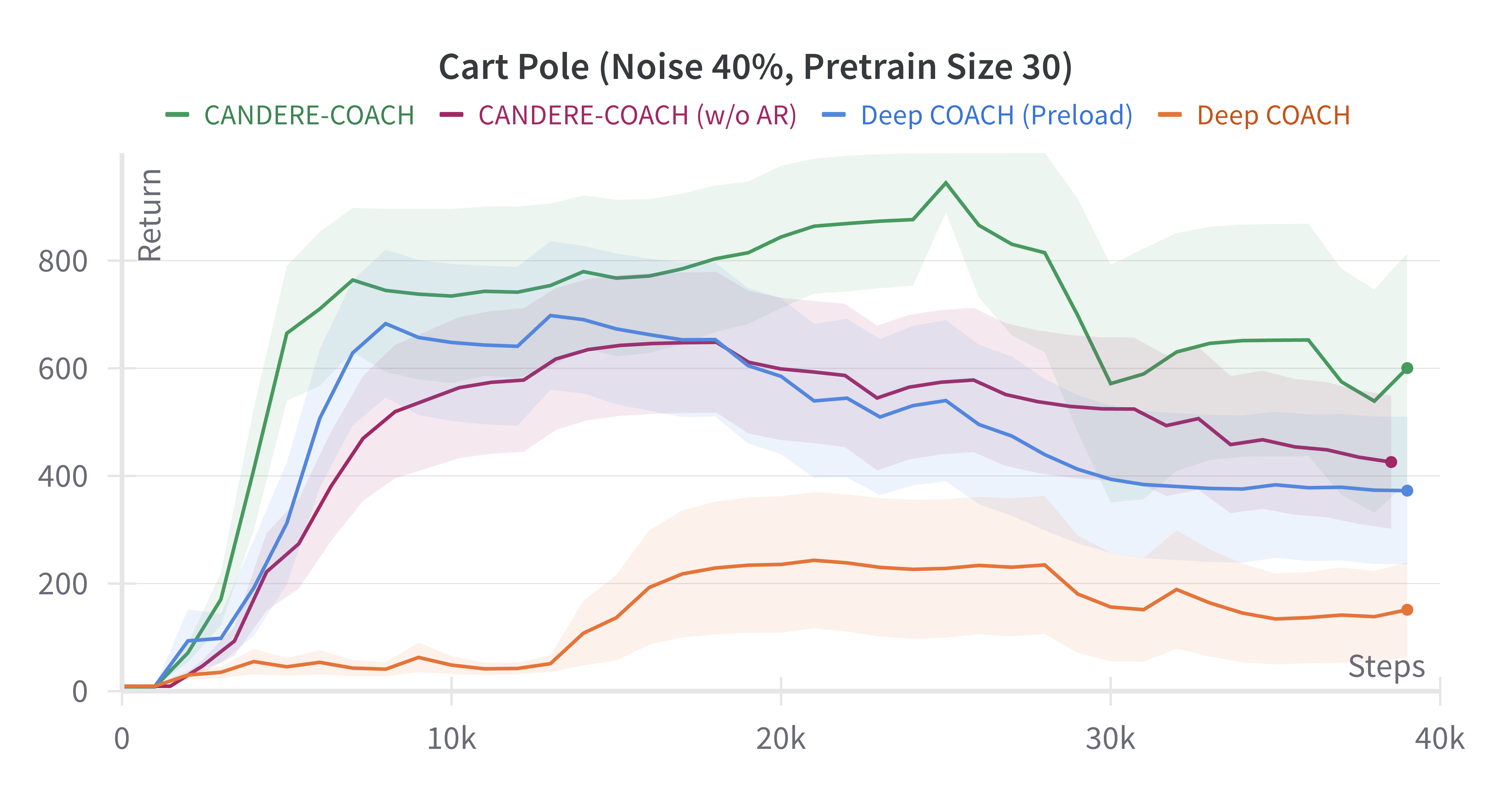}}
      \subfigure[]{\label{fig:all_plot_dk_n40}
      \includegraphics[width=0.32\textwidth]{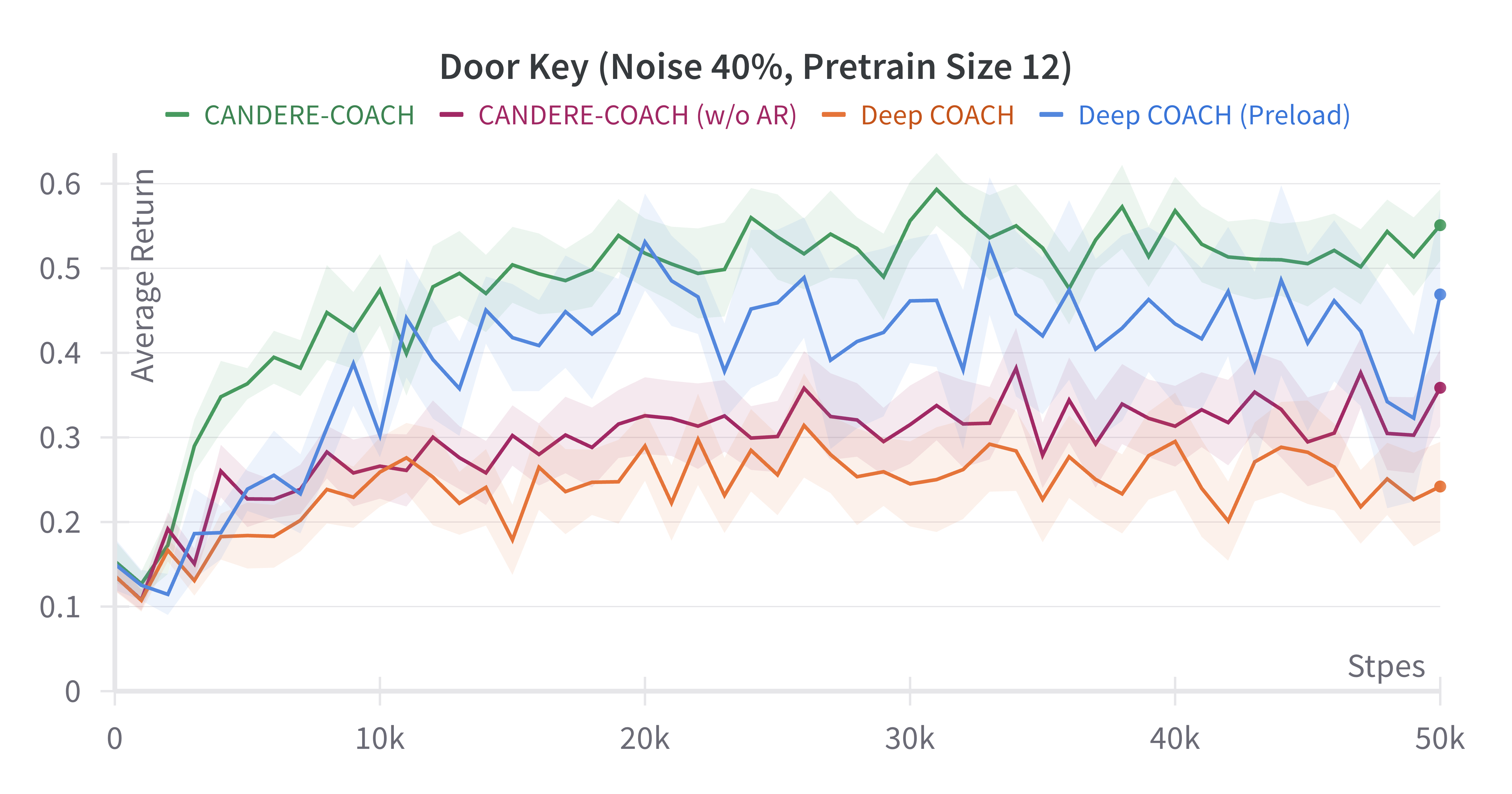}}
      \subfigure[]{\label{fig:all_plot_ll_n40}
      \includegraphics[width=0.32\textwidth]{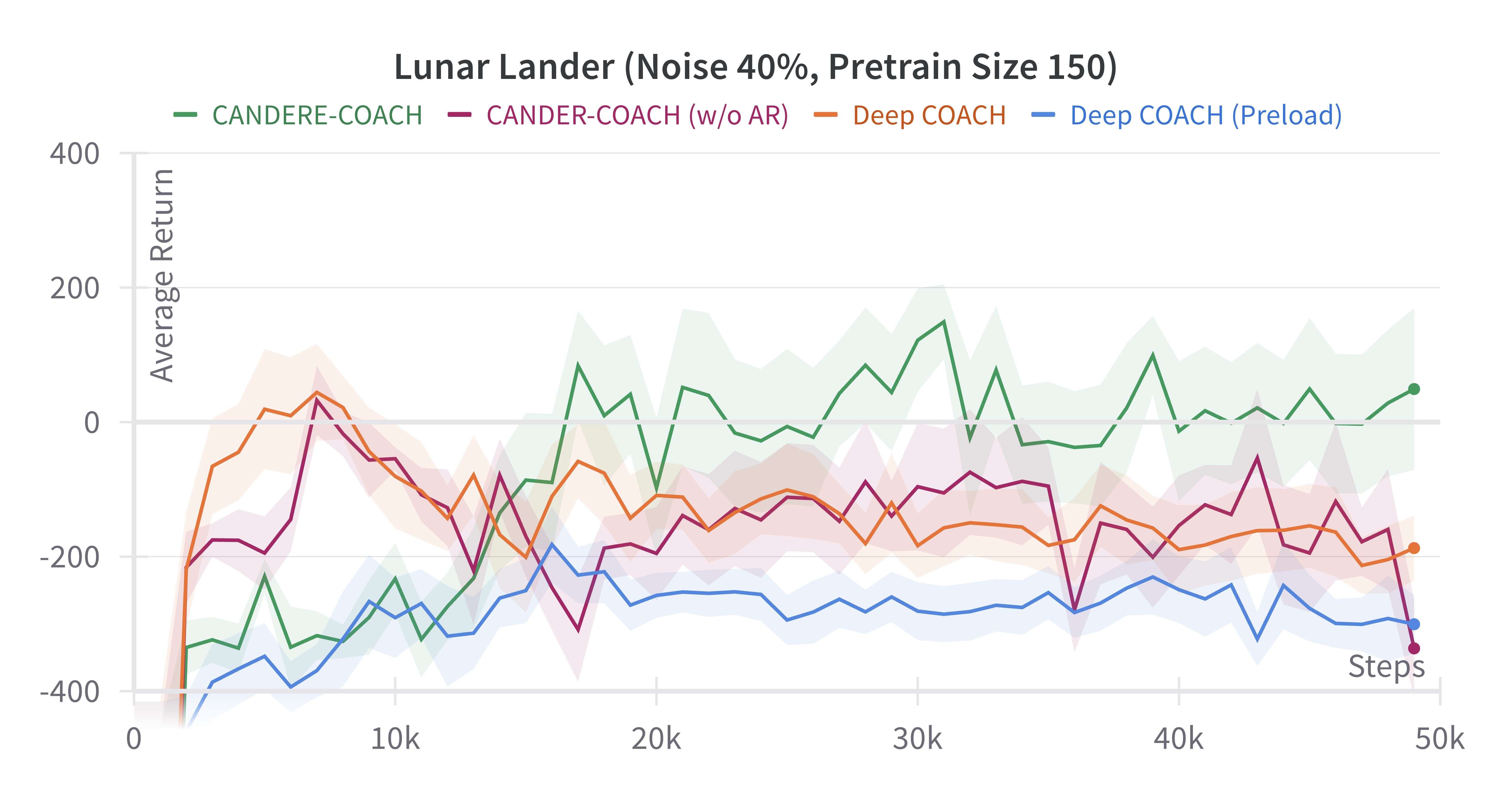}}
    
    \caption{Performance of CANDERE-COACH in Cart Pole, Door Key and Lunar Lander in 30\% and 40\% noise} 
    \label{fig:all_plot}
\end{figure*}

\begin{figure*}[h]
  \centering

    \subfigure[Average return in 10\% noise.]{
      \includegraphics[width=0.31\textwidth]{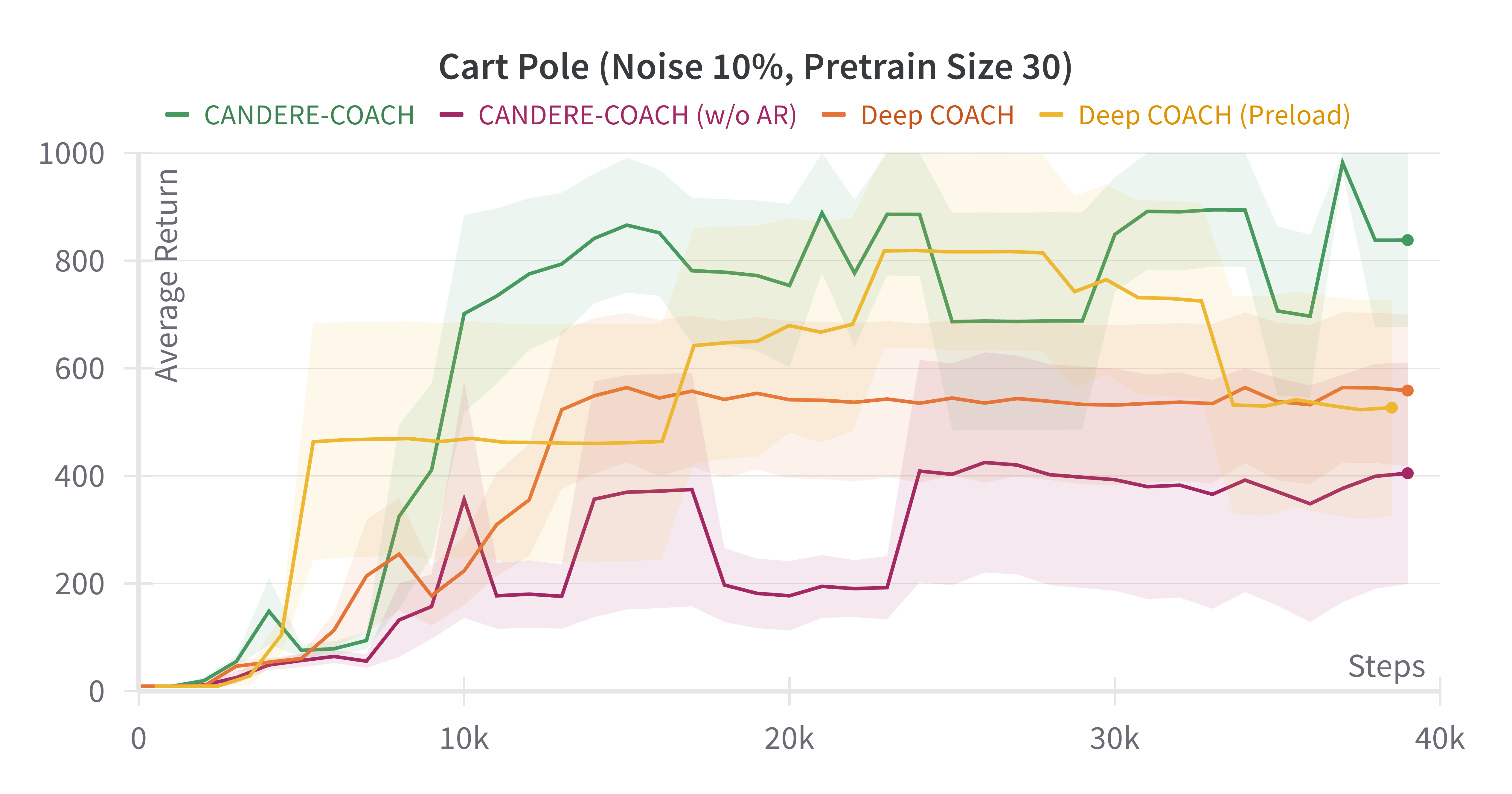}
        \label{fig:CANDERE_noisy_dataset_n10}
  }
  \subfigure[Average return in 20\% noise.]{
      \includegraphics[width=0.31\textwidth]{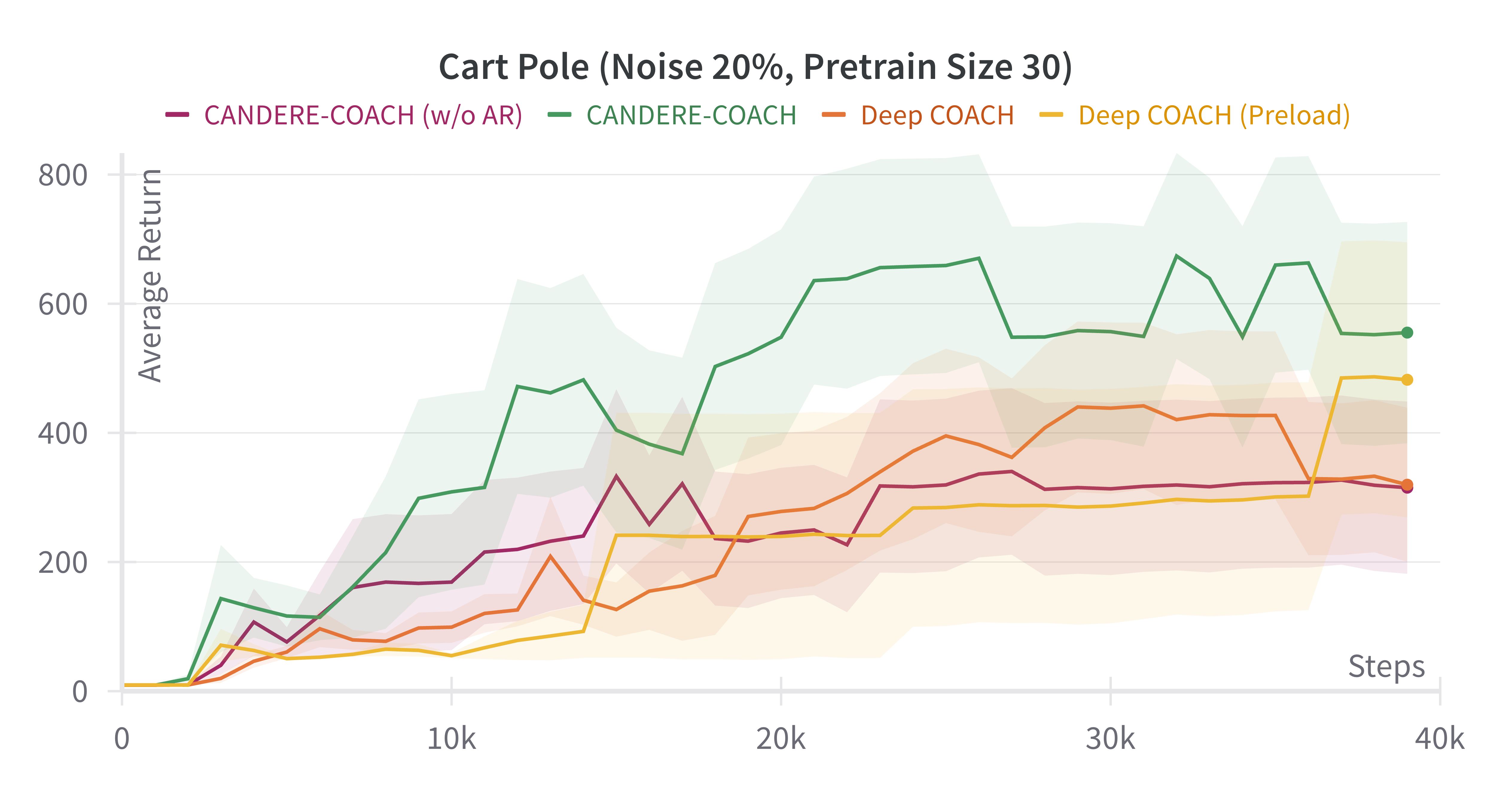}
        \label{fig:CANDERE_noisy_dataset_n20}
  }
  \subfigure[Average return in 30\% noise.]{
      \includegraphics[width=0.31\textwidth]{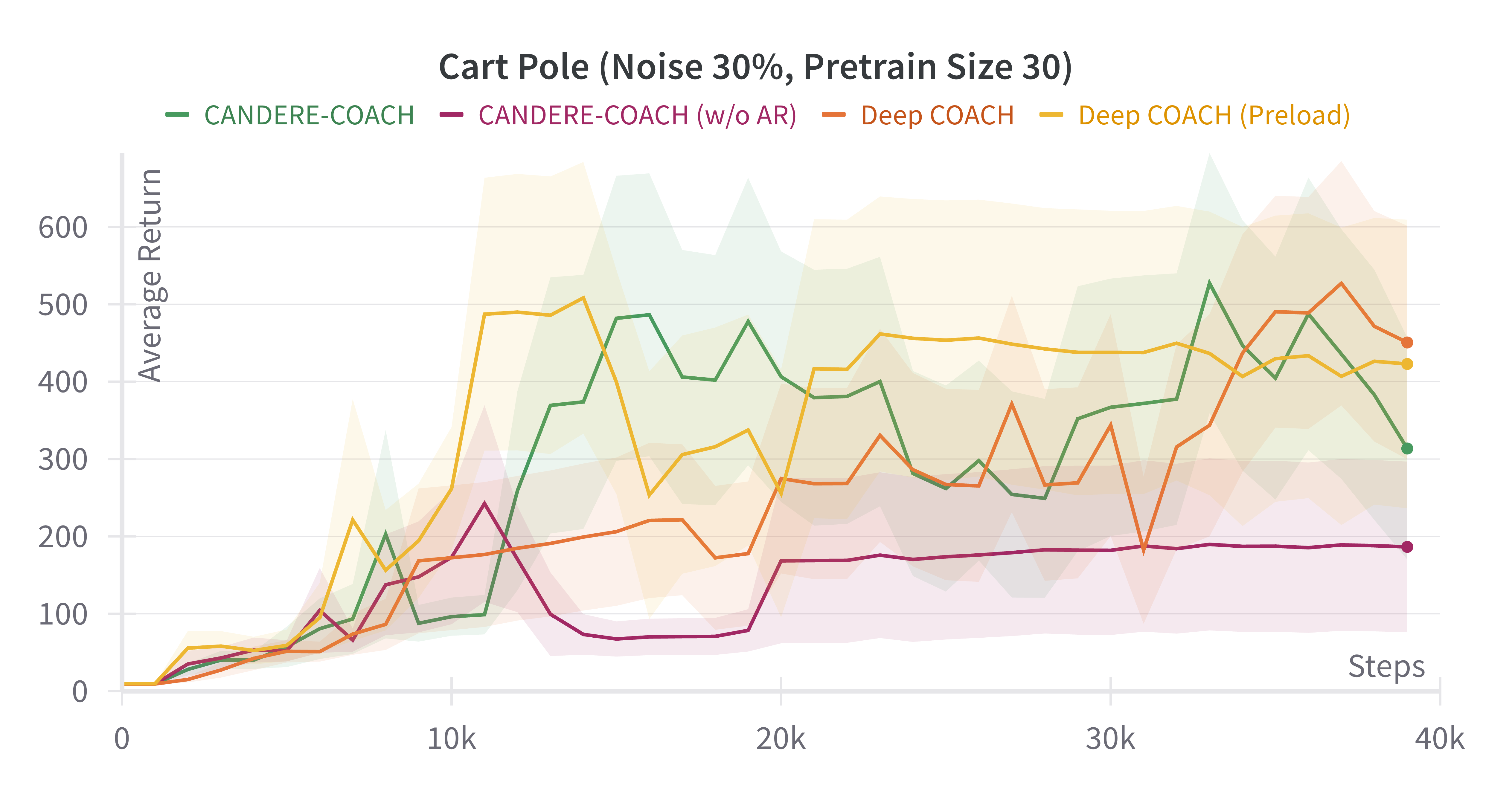}
        \label{fig:CANDERE_noisy_dataset_n30}
  }

\caption{Performance of \alg in Cart Pole, with noisy pretraining dataset}
\label{fig:CANDERE_noisy_dataset}
\end{figure*}

\noindent{\bf Cart Pole:} As shown in Figure~\ref{fig:all_plot_cp_n30}, CANDERE-COACH, can learn well (higher return) with 30\% noise in \textit{Cart Pole} and outperform Deep COACH, which shows highly unstable performance. CANDERE-COACH (w/o AR) is also able to outperform Deep COACH in performance, however, it cannot match CANDERE-COACH in performance which shows that active-relabeling for noise correction helps in identifying clean feedback from suspected noisy feedback.
With 40\% noise (shown in Figure~\ref{fig:all_plot_cp_n40}), CANDERE-COACH learns well, but the performance is unstable due to high noise.
However, it still outperforms CANDERE-COACH (w/o AR) 
in terms of average return towards the end of learning. Deep COACH fails to learn and receives low episodic rewards and a similar pattern is observed for Deep COACH (Preload).
\\
\noindent{\bf Door Key:}
As shown in Figure~\ref{fig:all_plot_dk_n30}, CANDERE-COACH is able to outperform Deep COACH and Deep COACH (Preload) under 30\% noise. 
CANDERE-COACH (w/o AR) does not perform well with almost the same performance as the baselines.
When the noise scale increases to 40\%, CANDERE-COACH (w/o AR) cannot perform well and only exceeds Deep COACH slightly in performance; though not statistically significant. 
The best performance is achieved by CANDERE-COACH as shown in Figure~\ref{fig:all_plot_dk_n40}, surpassing the baselines.\\
\noindent{\bf Lunar Lander:}
As suggested by Figure~\ref{fig:all_plot_ll_n30}, CANDERE-COACH (w/o AR) is not able to learn in this domain.
Although both baselines fail to achieve satisfactory performance, CANDERE-COACH (w/o AR) is impacted even more by noise, due to its classifier also receiving negative influence from noise during training and further deteriorates the learning ability of the agent.
CANDERE-COACH however, shows superior performance against 30\% noise, outperforming all the baselines that fail to reach an episodic return of 200. 
\footnote{In Lunar Lander, only an episodic return over 200 is considered a success.}
However, Lunar lander is still challenging under extremely high noise~($40\%)$ --- as shown in Figure~\ref{fig:all_plot_ll_n40},
CANDERE-COACH still achieves the best episodic return as compared to the baselines, but it is impacted by noise and fails to get an average return of 200 at the end.

In summary, CANDERE-COACH is generally effective in filtering noise compared to other baselines. Its performance on average is better with relatively less noise (up to $30\%$), as compared to very high noise ($40\%$); thus affirmatively answering RQ2. CANDERE-COACH with active relabeling is always statistically significantly better 
in performance as compared to all the other algorithms, thus supporting RQ3. 
\footnote{For the performance of CANDERE-COACH with other noise levels, please refer to Section~\ref{sec:candere_coach_other_noise} of the Appendix. }

\subsection{Ablation studies}
\label{sec:abaltion_studies}
\begin{figure}[h]
  \centering
    \includegraphics[width=0.45\textwidth]{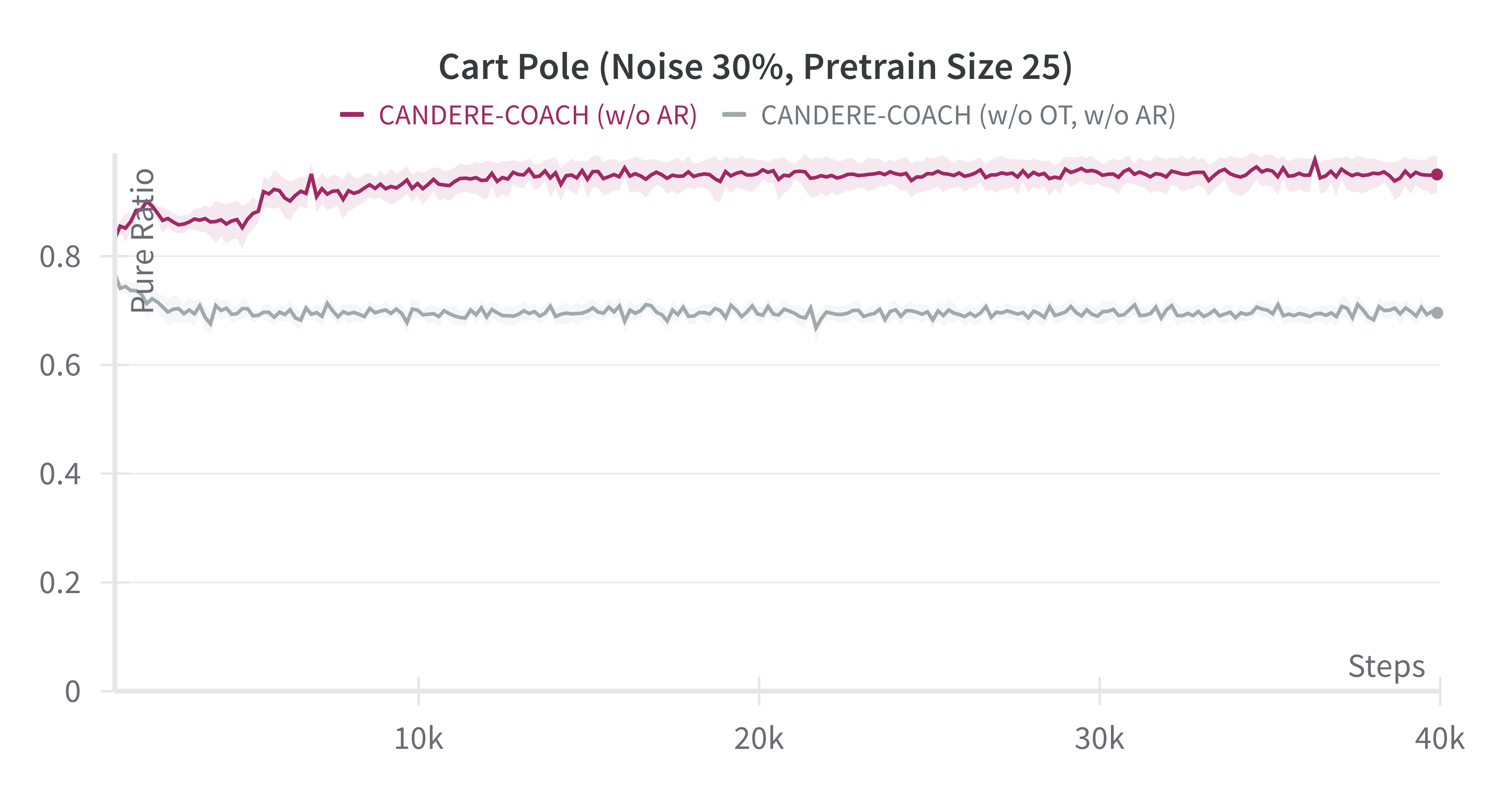}
    
  \caption{Pure ratio with and without online training in CartPole in 30\% noise. While the agent explore new states and actions, the distribution of state and action changes, and therefore a fixed classifier is predicting less accurately over time without online training.}
    \label{fig:abaltion_ot_n30_pr} 
\end{figure}

We conduct additional ablation studies to understand the components and effects of hyperparameters on the performance of CANDERE-COACH.\\
\noindent \textbf{Effect of online training on noise-filtering classifier:} 
Online training of the classifier tries to balance two problems. First, the state action distribution can shift (relative to the data used to pretrain the classifier), suggesting online training will help. Second, trying to update the classifier with noisy labels could hurt performance, suggesting online training will not help.
We denote the CANDERE-COACH without online training and active relabelling as CANDERE-COACH (w/o AR, w/o OT).
Figure~\ref{fig:abaltion_ot_n30_pr} shows that for CANDERE-COACH (w/o AR, w/o OT), the pure ratio reduces over time, suggesting that as the agent explores different areas of the state-action space, the distribution shift leads to worse classifier performance. 
In contrast, online training allows the pure ratio to gradually increase.
Eventually, the pure ratio stabilizes around 95\% 
and as a result, \alg (w/o AR) is able to learn robustly against 30\% noise with pretraining dataset of size 25.
\footnote{Plot of average return in 30\%, as well as results in 40\% can be found in Appendix (Section~\ref{sec:abaltion_study_online_training}).}
To summarise,
with online training, \alg has a better sample efficiency and leads to a higher pure ratio during training.

\noindent \textbf {Effect of noisy pretraining Dataset:}
\begin{figure*}[h]
  \centering
  \subfigure[Average return in 30\% Noise]
  {
  \label{fig:candere_tamer_n30}
  \includegraphics[width=0.45\textwidth]{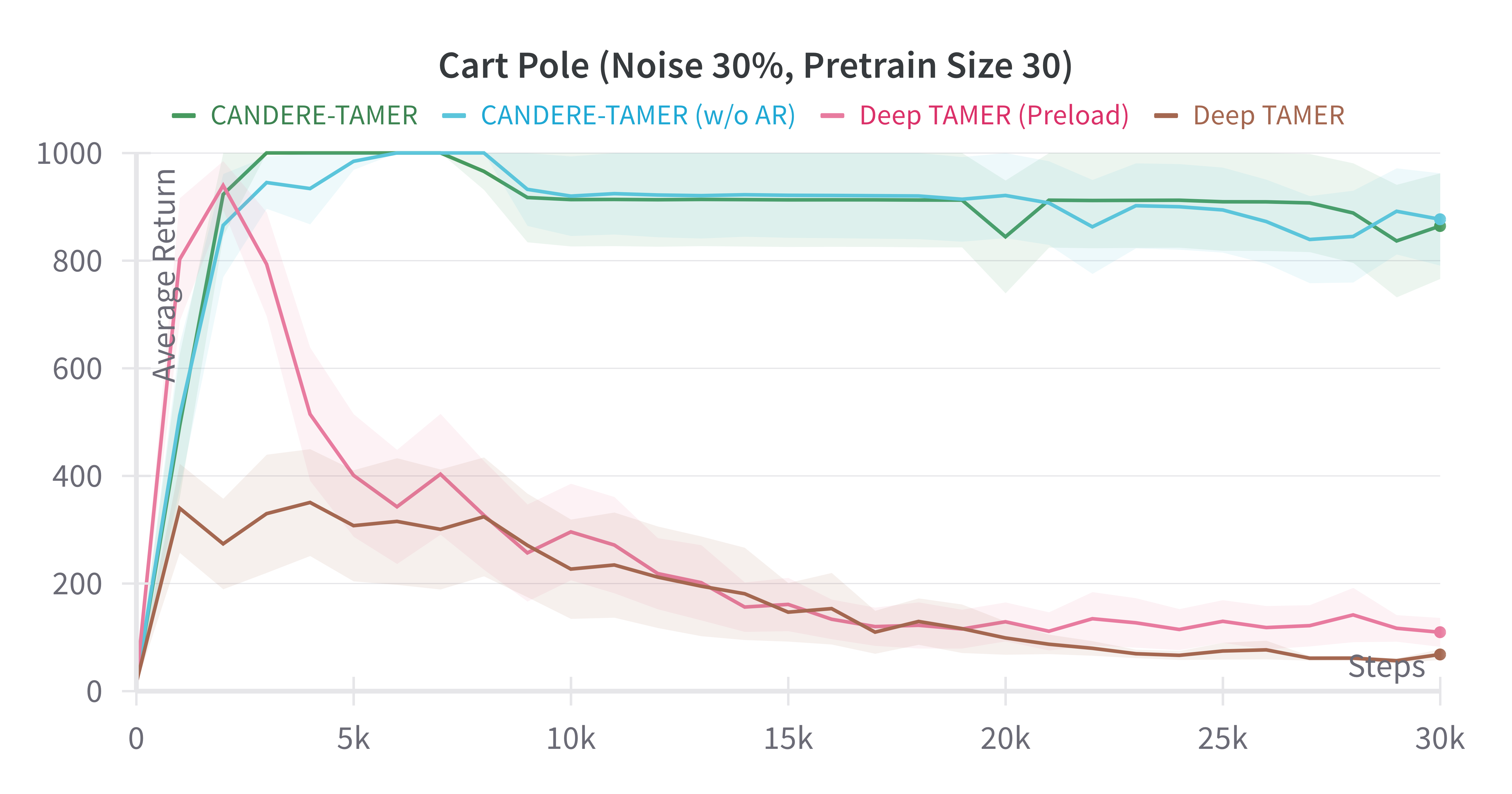}
  }
  \subfigure[Average return in 40\% Noise]{\label{fig:candere_tamer_n40}
  \includegraphics[width=0.45\textwidth]{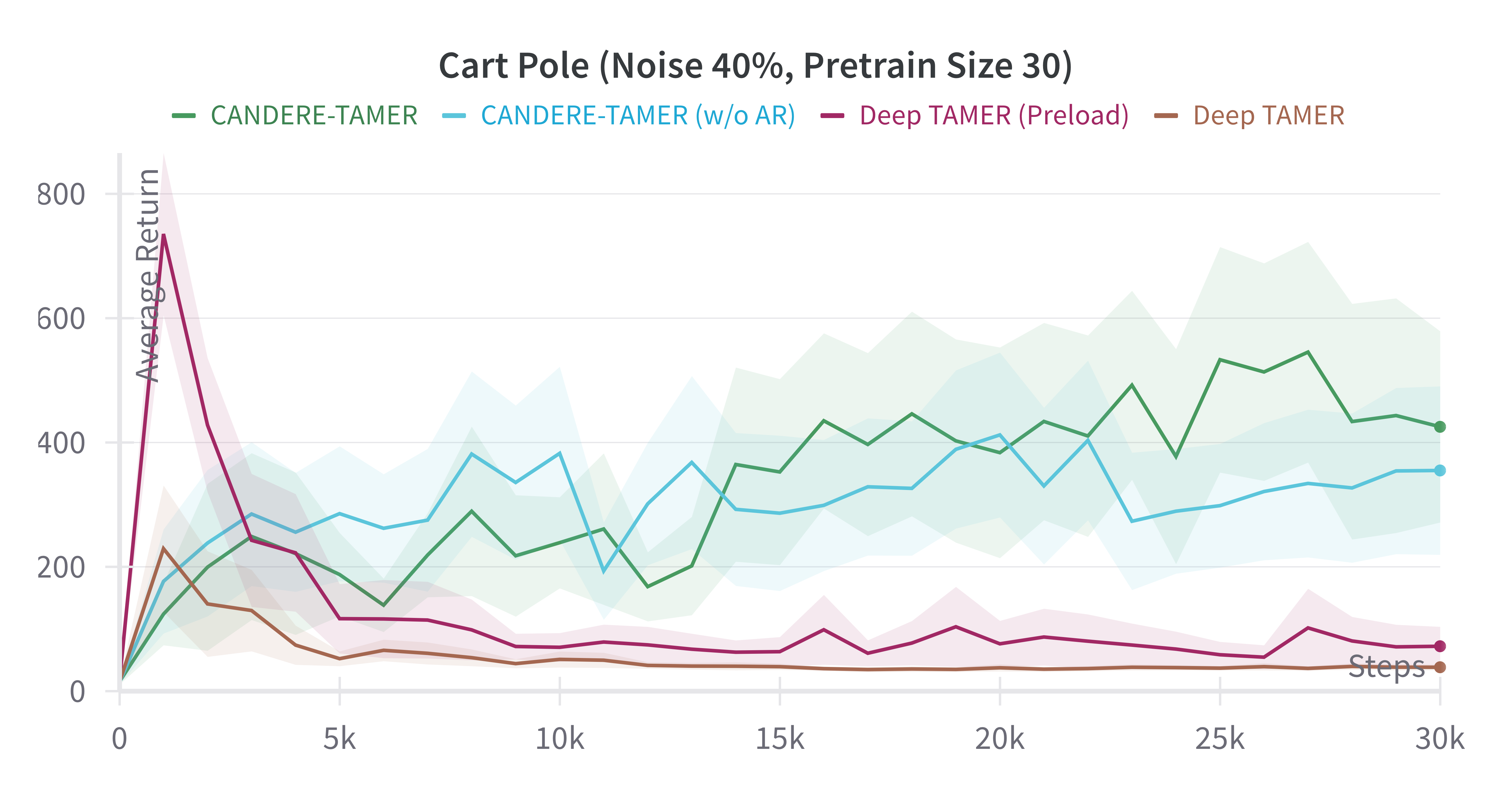}}
\caption{Performance comparison of CANDERE-TAMER in Cart Pole}
\label{fig:candere_tamaer}
\end{figure*}
In prior sections, CANDRE-COACH was allowed access to a small noise-free feedback dataset for pretraining the classifier. In this section, we invalidate this assumption by testing CANDERE-COACH with a noisy pretraining dataset. 
As observed in Figure~\ref{fig:CANDERE_noisy_dataset},  CANDERE-COACH can still perform well under low amounts of noise, such as 10\% and 20\% (see Figures~\ref{fig:CANDERE_noisy_dataset_n10} and Figure~\ref{fig:CANDERE_noisy_dataset_n20}), while CANDERE-COACH (w/o AR) cannot outperform our baseline.
However, if the noise level is too high (30\%), as shown in Figure~\ref{fig:CANDERE_noisy_dataset_n30}, we observe that the performance of CANDERE-COACH begins to degrade.
Also, pretraining with a noisy dataset reduces the overall performance compared to the performance reported in previous sections, where a noise-free pretraining dataset is available. 
To summarise, CANDERE-COACH can work with noisy pretraining dataset but it only shows promising results under reasonably low noise levels like 10\% or 20\% noise.

\subsection{Extending to CANDERE-TAMER}

There is no fundamental challenge to expand our proposed de-noising mechanism to other learning from feedback algorithms. 
Here, we present CANDERE-TAMER, which is similar to CANDERE-COACH but built based on Deep TAMER~\cite{warnell2018deep}.
The results are shown in Figure~\ref{fig:candere_tamaer}. 
We observe a similar pattern in which our algorithms outperform our baselines (Deep TAMER and Deep TAMER (Preload)) and learn successfully against  30\% noise, while the performance of baselines degrades poorly with noise. In 40\% noise, CANDERE-TAMER still shows a better average return and outperforms our baselines; however, its performance is significantly worse than that with $30\%$ noise. This experiment shows the potential of our approach to be used as a plug and play tool inside any human-in-the-loop RL algorithm, which will be explored as future work.

\section{Conclusion and Future Work}
In this paper, a new algorithm is proposed inside the learning from feedback RL framework.
Through experiments in multiple settings and tasks, we show that the \alg is able to handle up to 40\% noise, with a small noise-free feedback dataset, outperforming the baselines.
We also show that if the noise is small enough, our \alg can also work with a noisy pretraining dataset.
Besides Deep COACH, we further show that the proposed noise detection mechanism can be extended to Deep TAMER.
For future work, we intend to expand this framework to learning from preferences algorithms and also test its performance with different types of noise, such as non-symmetric and feature-dependent noise, as well as conducting a human subject study.

\begin{ack}
Part of this work has taken place in the Intelligent Robot Learning (IRL) Lab at the University of Alberta, which is supported in part by research grants from the Alberta Machine Intelligence Institute (Amii); a Canada CIFAR AI Chair, Amii; Digital Research Alliance of Canada; Huawei; Mitacs; and NSERC.
\end{ack}

\bibliography{neurips_2024}

\clearpage

\appendix

\section{Deep COACH with noise}
\label{sec:deep_coach_with_noise}
\begin{figure}[h]
  \centering
  \subfigure[Unlimited budget of feedback ]
  {\label{fig:coachdifferentnoise_dk}
  \includegraphics[width=0.5\textwidth]{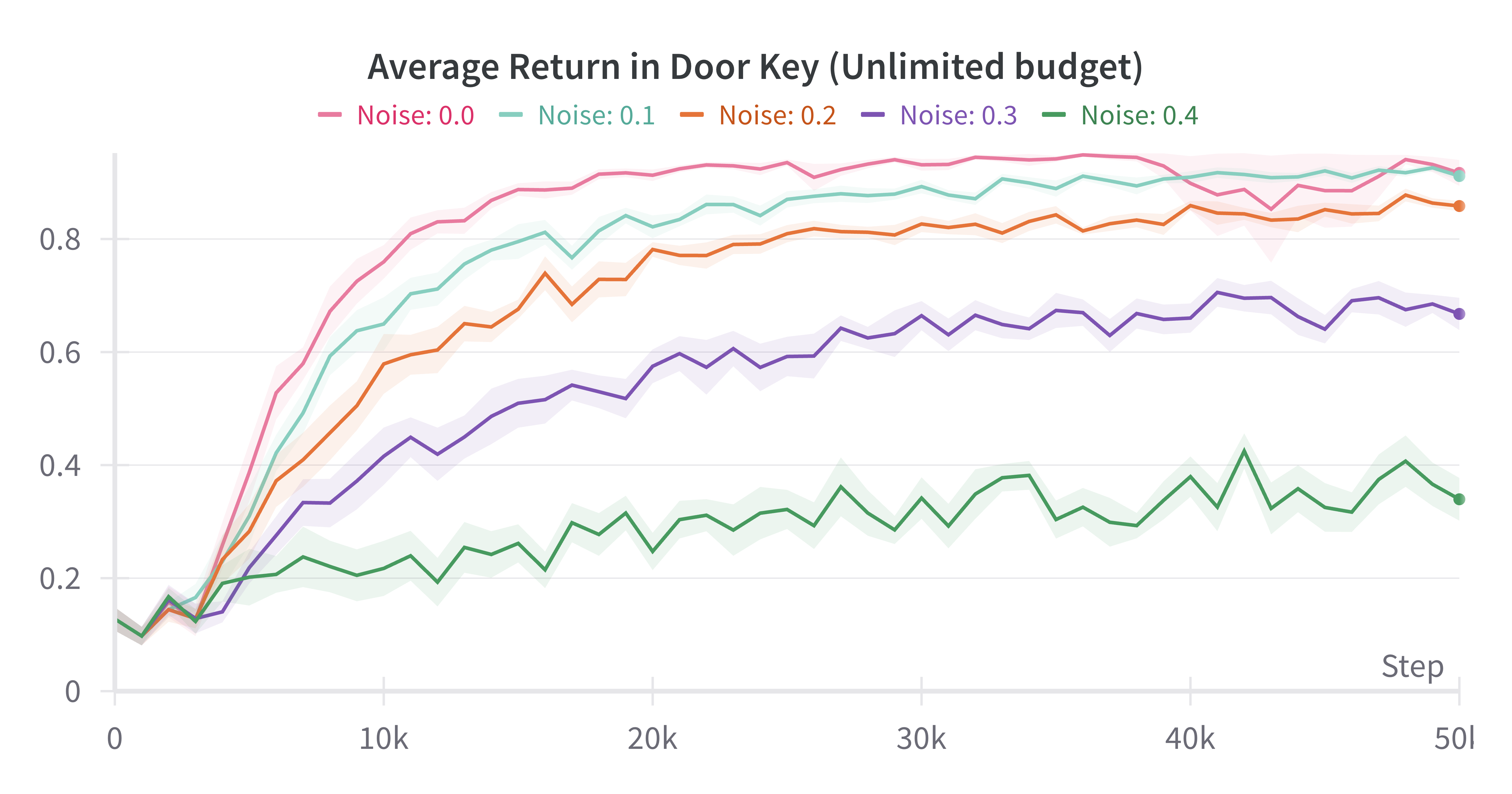}}
  \vfill
  \subfigure[Limited budget (500) of feedback]{\label{fig:coachdifferentnoiselimited_budget_dk}
  \includegraphics[width=0.5\textwidth]{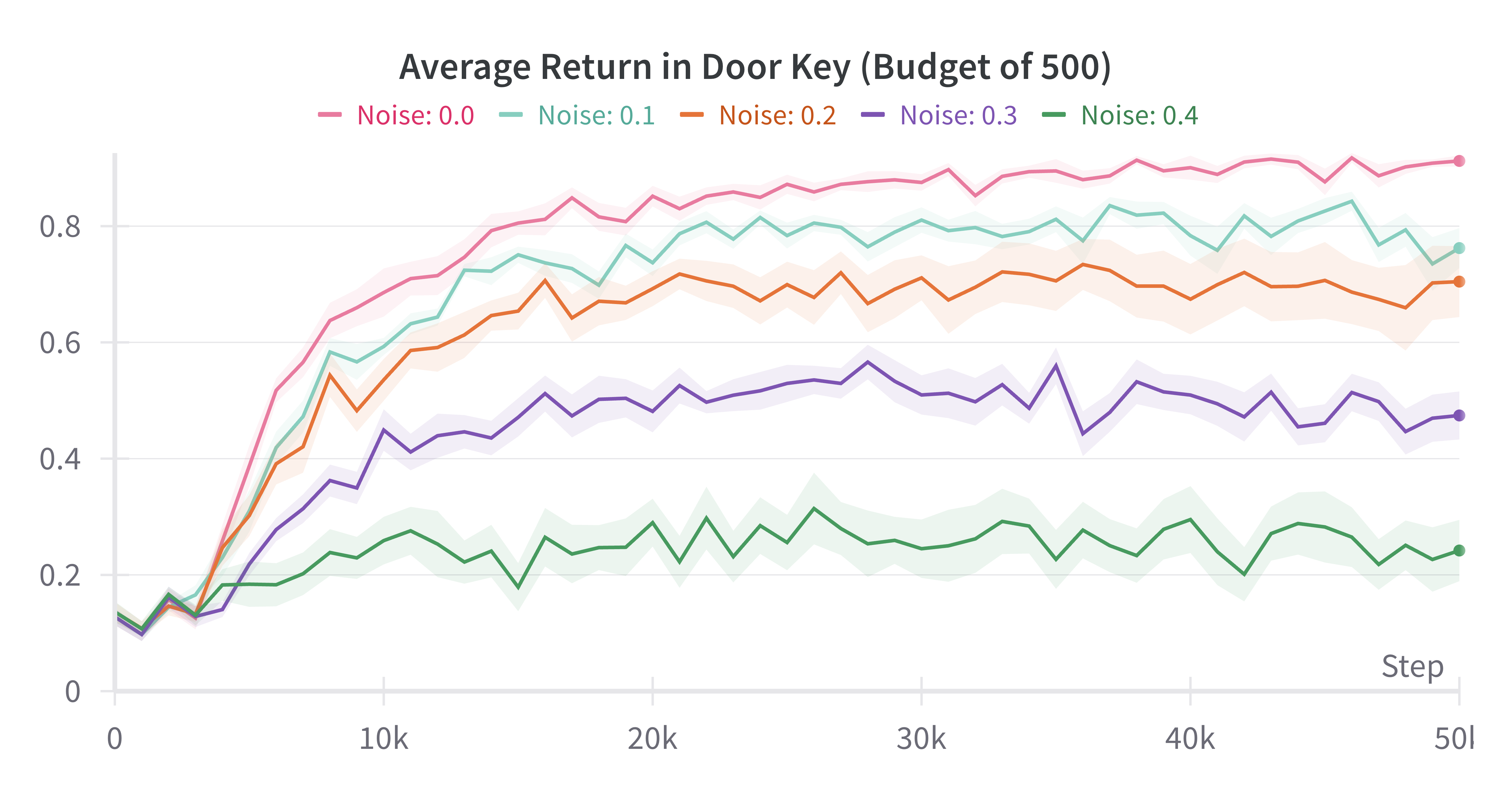}}
\caption{Performance of Deep COACH under different scales of noises in Door Key }
\label{fig:exp1_dk}
\end{figure}

We also conducted experiments to study the impact of noise on our baseline (Deep COACH) in other domains.
In Door Key, noticeably, the performance of COACH seems less impacted with a limited budget with a smaller average return drop, shown in Figure~\ref{fig:exp1_dk}.
This is mainly due to the unique settings in Door Key. 
In Door Key, the agent needs to grab the key, unlock the door, and reach the destination. 
An evaluative reward of positive +1 discounted by time steps is given when it reaches the destination.
Even a randomly initialised agent can reach the destination and achieve a nonzero return at step 0, which is different from Cart Pole and Lunar Lander, where a bad policy results in low episodic returns. 
Therefore, the performance in Door Key seems to receive less negative impact with a limited budget. 
However, we still observe a drop in performance. 
For example, in 40\% noise, with an unlimited feedback budget, the return reaches up to 0.4 after 50k steps, while with a limited budget it converges around 0.2. 

\begin{figure}[h]
  \centering
  \subfigure[Unlimited budget of feedback ]
  {\label{fig:coachdifferentnoise_ll}
  \includegraphics[width=0.5\textwidth]{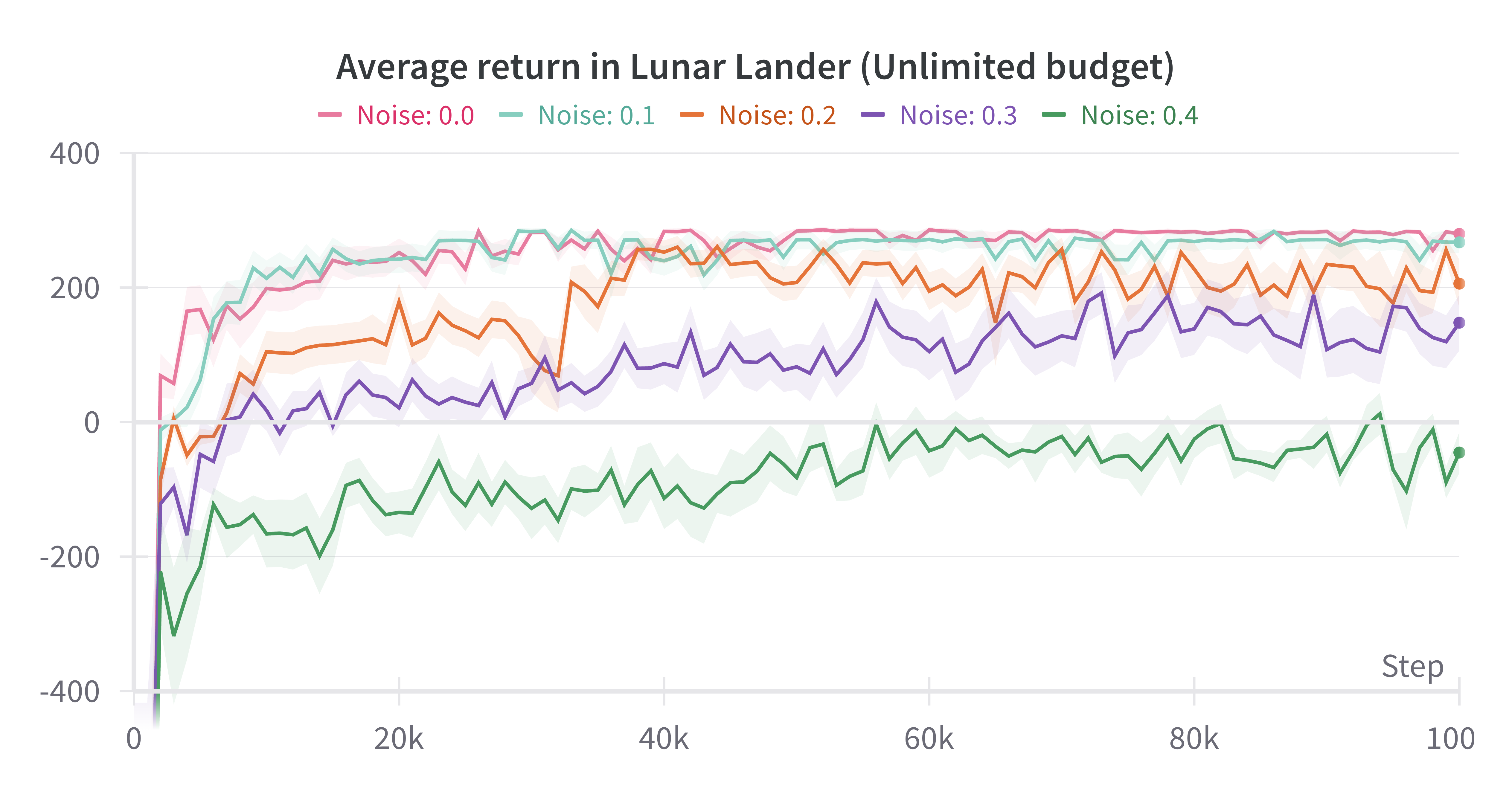}}
  \vfill
  \subfigure[Limited budget (2000) of feedback]{\label{fig:coachdifferentnoiselimited_budget_ll}
  \includegraphics[width=0.5\textwidth]{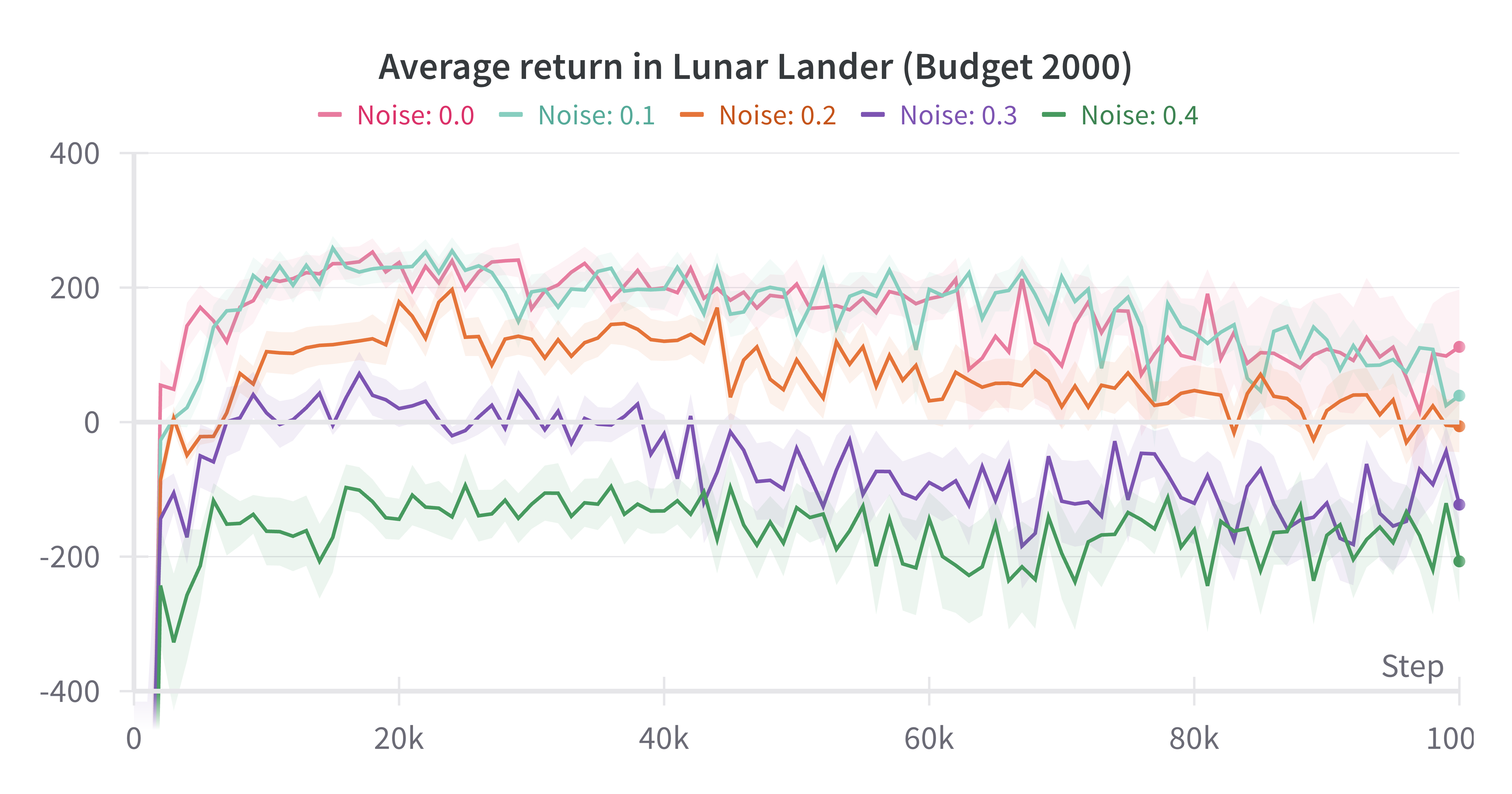}}
\caption{Performance of Deep COACH under different scales of noises in Lunar Lander}
\label{fig:exp1_ll}
\end{figure}

For Lunar Lander, we observe the same trends as Cartpole, as shown in Figure~\ref{fig:exp1_ll}; the agent performance for unlimited feedback setting decreases as noise $\%$ increases. However, the agent is still able to learn, though not as good as learning from clean feedback (0\% noise). With limited feedback, the agent's learning performance deteriorates significantly and the agent eventually unlearns over time.

\section{CANDERE-COACH with unlimited budget}
In this section, we evaluate the performance of the original \alg with an unlimited budget.
We conducted experiments with different sizes of pretraining datasets for the classifier.

\begin{figure}
  \centering
  \includegraphics[width=0.5\textwidth]{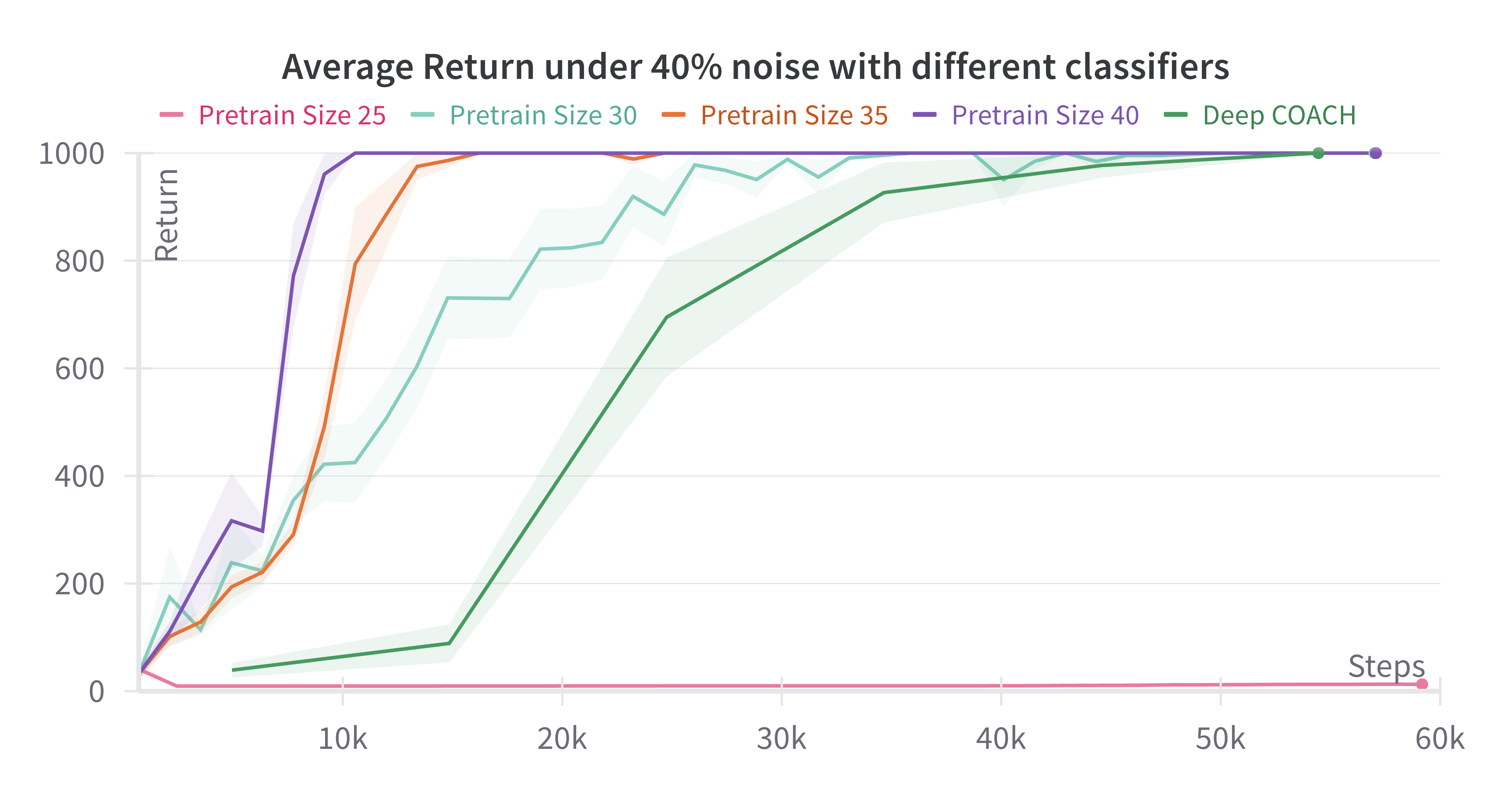}
\caption{Performance of \alg under 40\% noise, with different amounts of pretraining data size.}
\label{fig:exp2}
\end{figure}

As can be seen from Figure~\ref{fig:exp2}, \alg is able to outperform when we have a pretraining data size of 30, against 40\% of noise.
Though CANDERE is able to learn faster than the baseline, with an unlimited budget of feedback, both the COACH and CANDERE is able to learn and reach almost perfect performance eventually. 
Furthermore, when the classifier does not have enough data for pretraining and cannot select correct labels for agent updates, the performance will be downgraded to a very low level.
Here, to conclude, with a small amount of feedback dataset, \alg is able to outperform Deep COACH by learning faster. 
But with an unlimited budget of feedback, even our baseline is able to learn well against high noise. 

\section{More ablation study on online training}
\label{sec:abaltion_study_online_training}

\begin{figure*}[h]
  \centering
  \subfigure[Average return with and without online training in CartPole in 30\% noise.]{
      \includegraphics[width=0.45\textwidth]{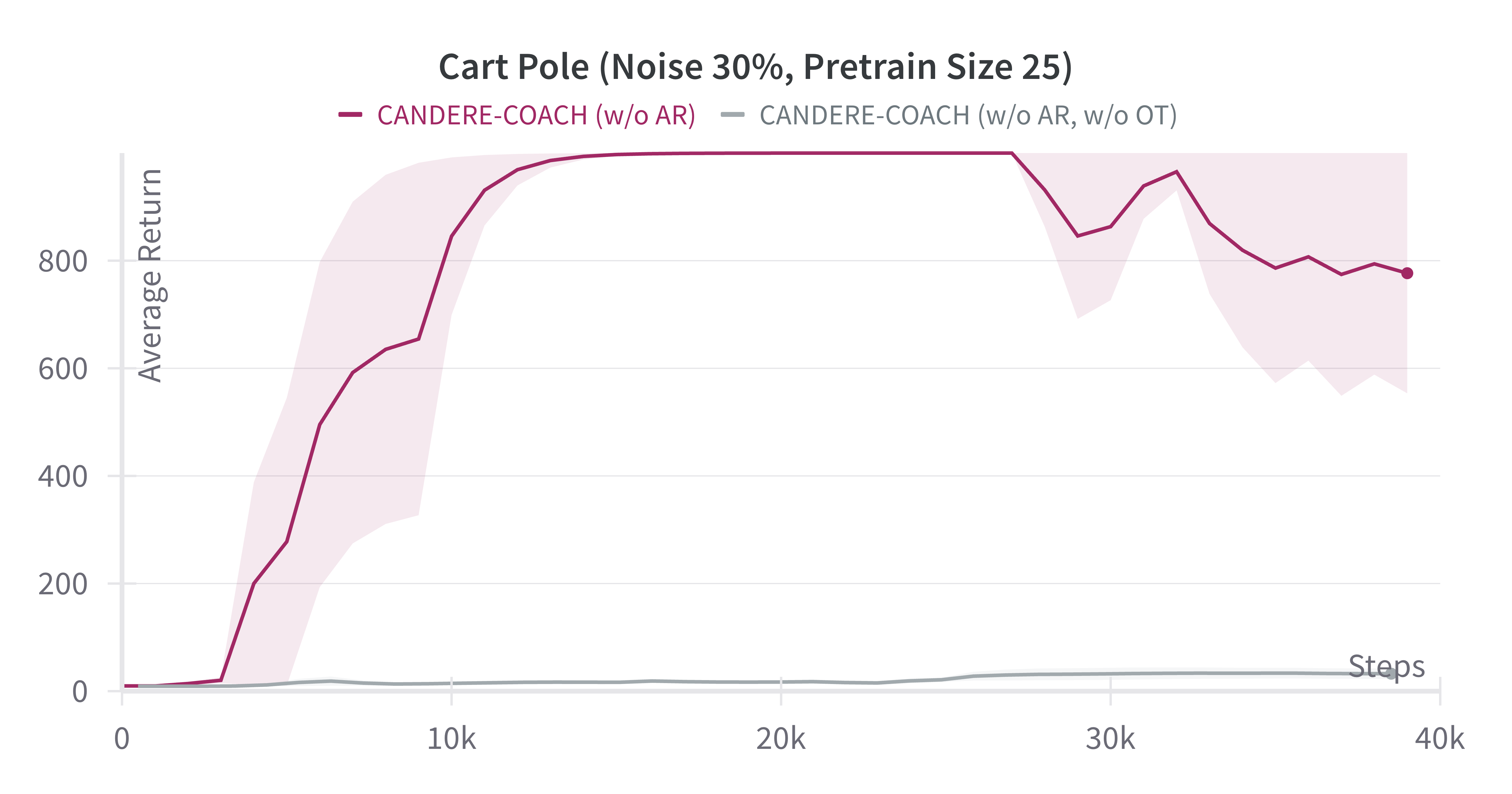}
        \label{fig:abaltion_ot_n30_ar}
  }
  
  \caption{With online learning, CANDERE-COACH (w/o AR) can succeed with only pretraining dataset of size 25.}
    \label{fig:ablation_ot_n30}
\end{figure*}

\begin{figure}[h]
  \centering
  \subfigure[Average return with and without online training in Cart Pole in 40\% noise.]{
      \includegraphics[width=0.45\textwidth]{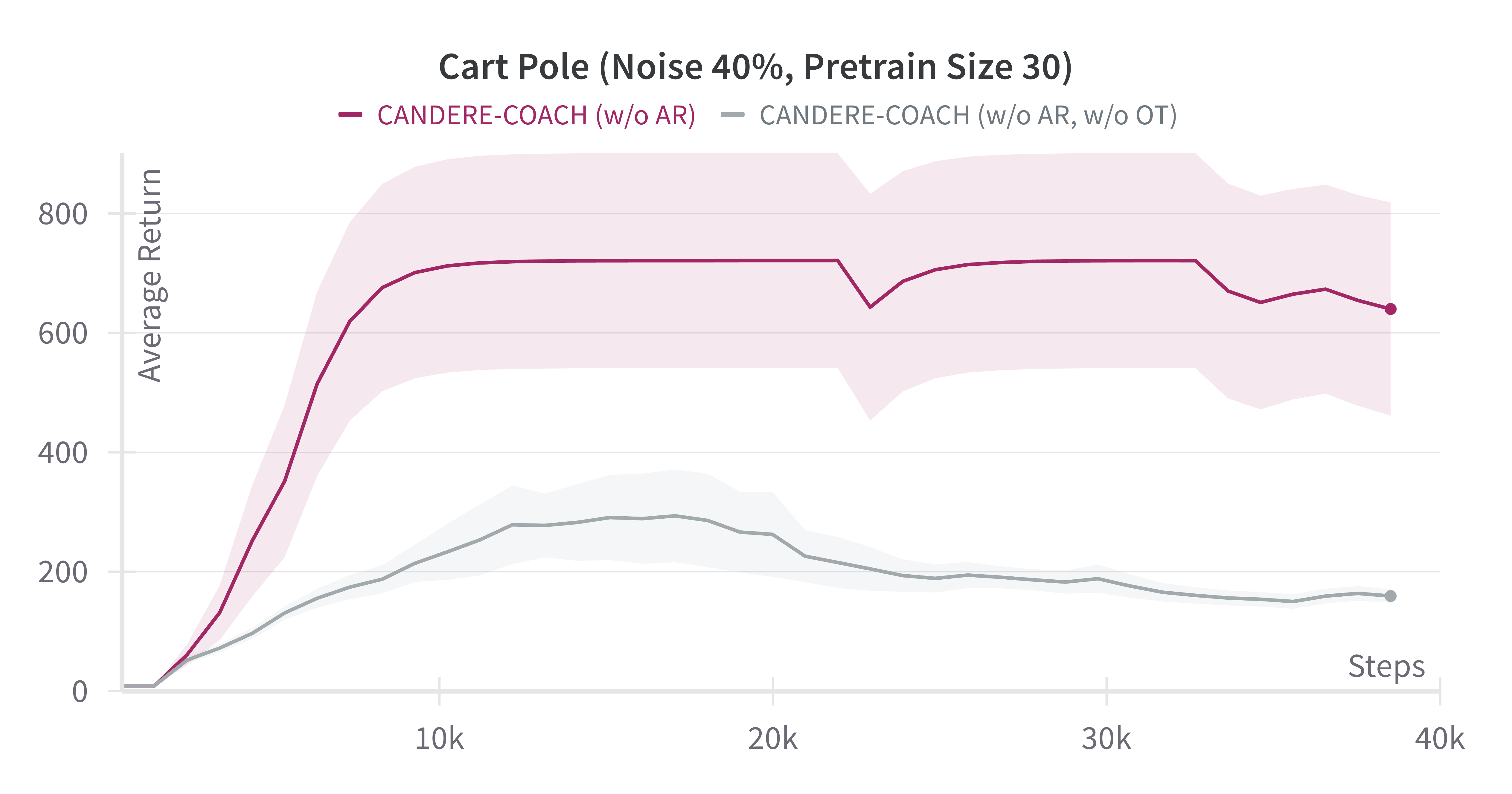}
        \label{fig:abaltion_ot_n40_ar}
  }
  \subfigure[Pure ratio with and without online training in Cart Pole in 40\% noise.]{
  \includegraphics[width=0.45\textwidth]{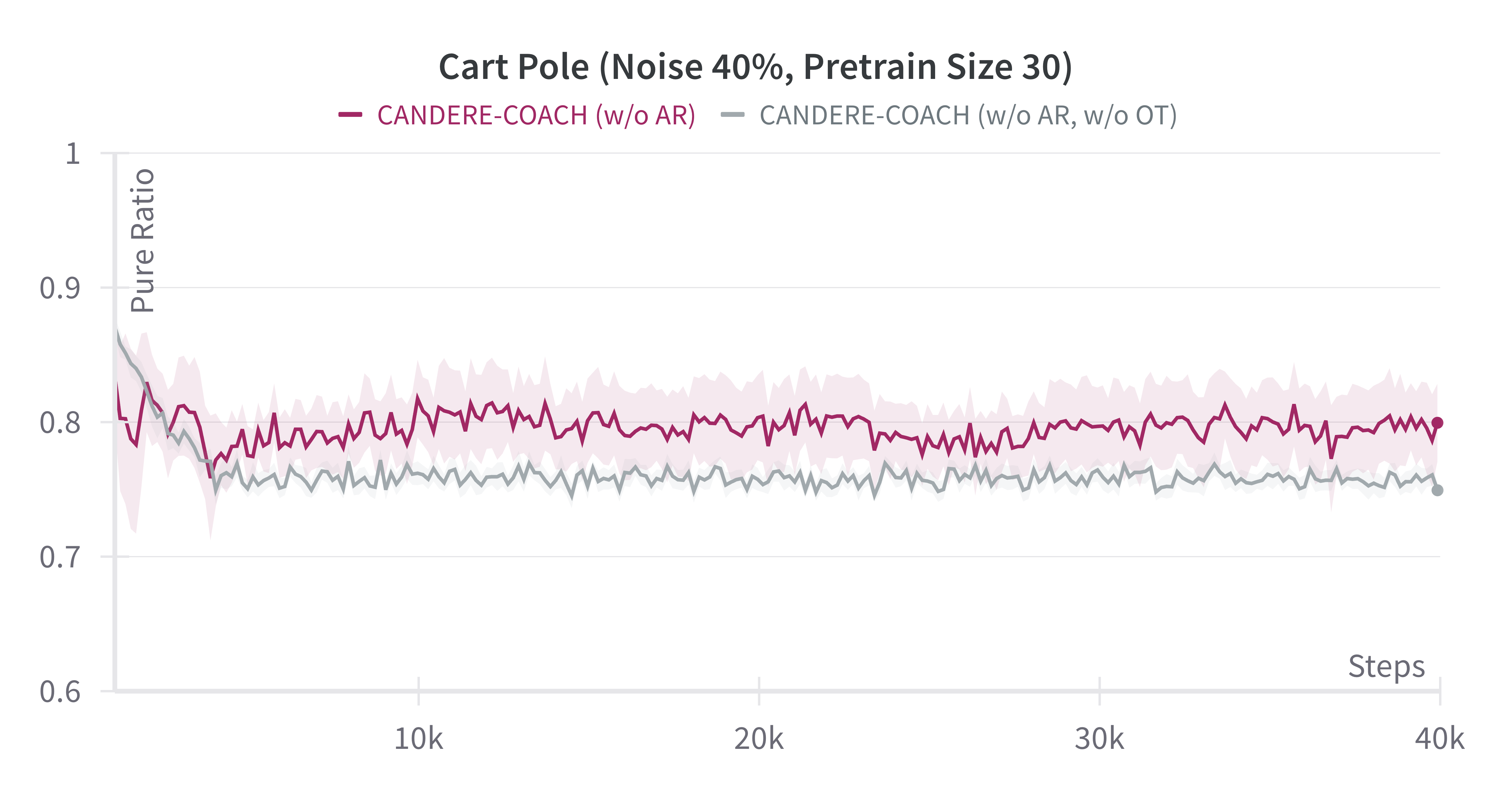}
\label{fig:abaltion_ot_n40_pr}  
}
  \caption{Ablation study on online training: average episode return and pure ratio of \alg (with and without OT) and Deep COACH in Cart Pole under 40\% noise.}
    \label{fig:ablation_ot_n40}
\end{figure}

Online training does not always show such a great improvement when the noise level increases.
The results with 40\% noise can be found in Figure~\ref{fig:ablation_ot_n40}.
We also observe that the pure ratio under extremely high noise is reduced and unstable, which differs from the pattern seen in 30\% noise, which shows that the online training mechanism tends to perform worse under very high noise.
However, its performance remains better than CANDERE-COACH without online training both in average return (shown in Figure~\ref{fig:abaltion_ot_n40_pr}) and pure ratio (shown in Figure~\ref{fig:abaltion_ot_n40_ar}). 

\section{Ablation study on pretraining sizes}
\begin{figure}
  \centering
  \includegraphics[width=0.45\textwidth]{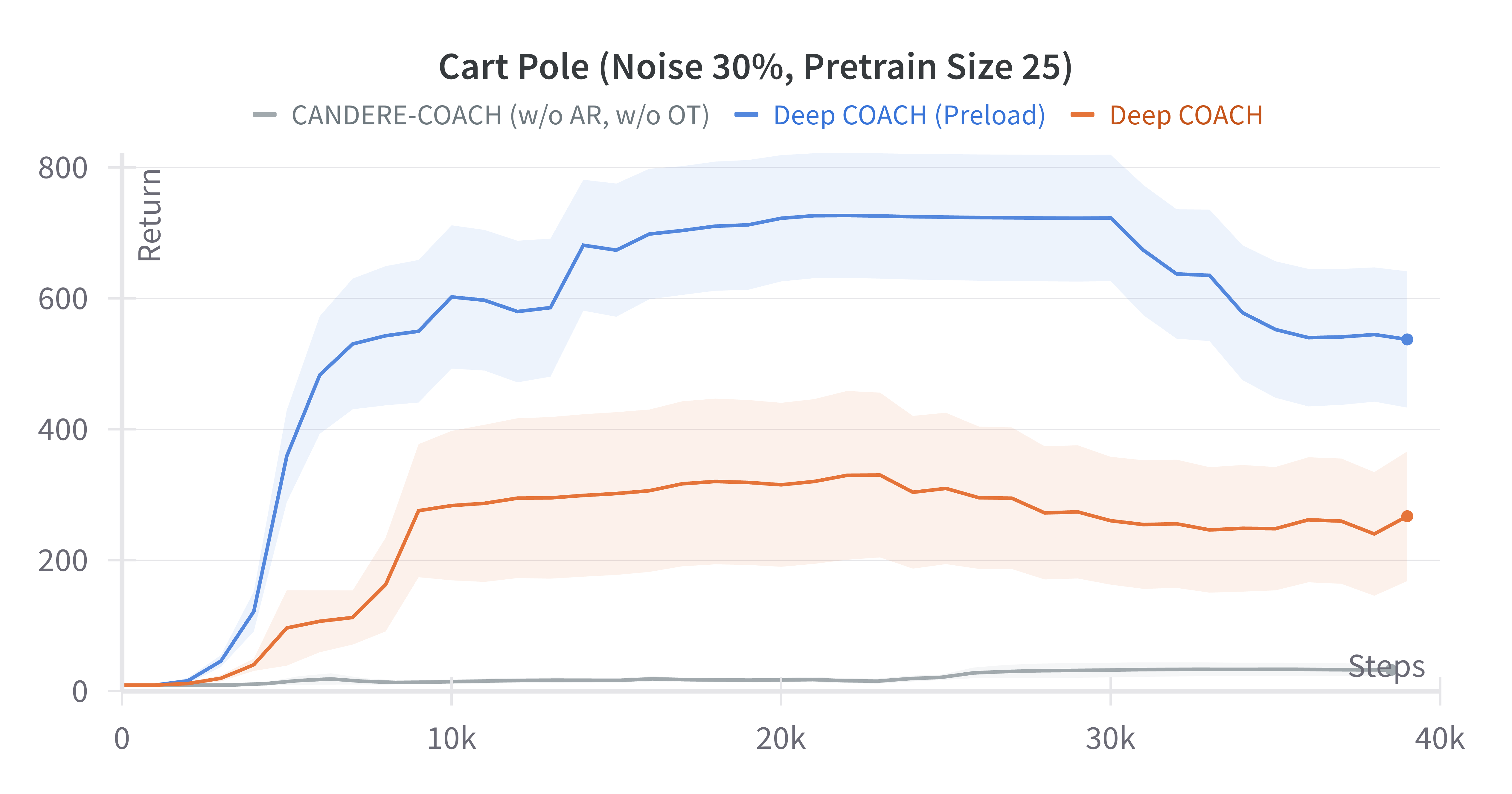}
  \includegraphics[width=0.45\textwidth]{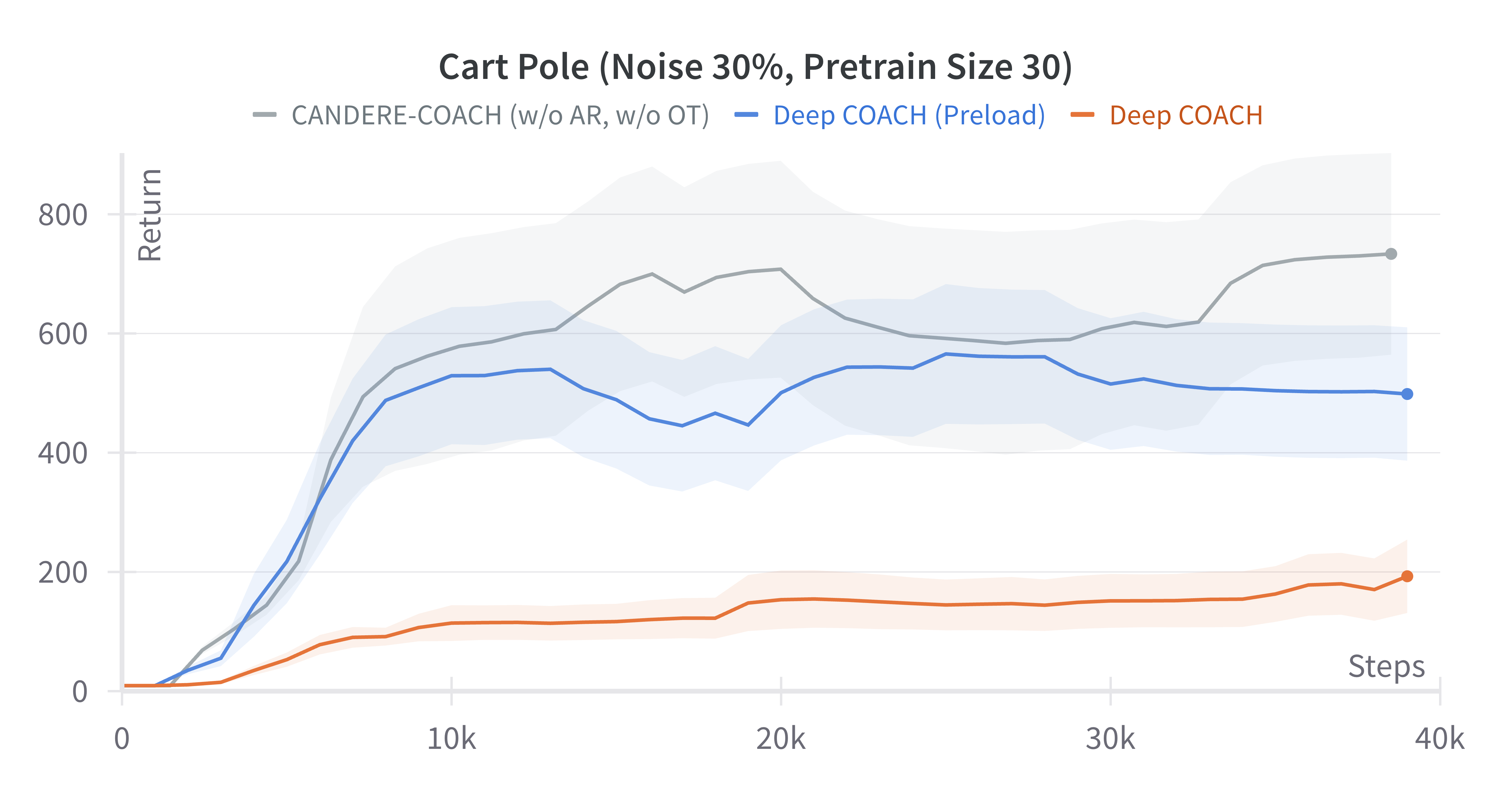}
  \includegraphics[width=0.45\textwidth]{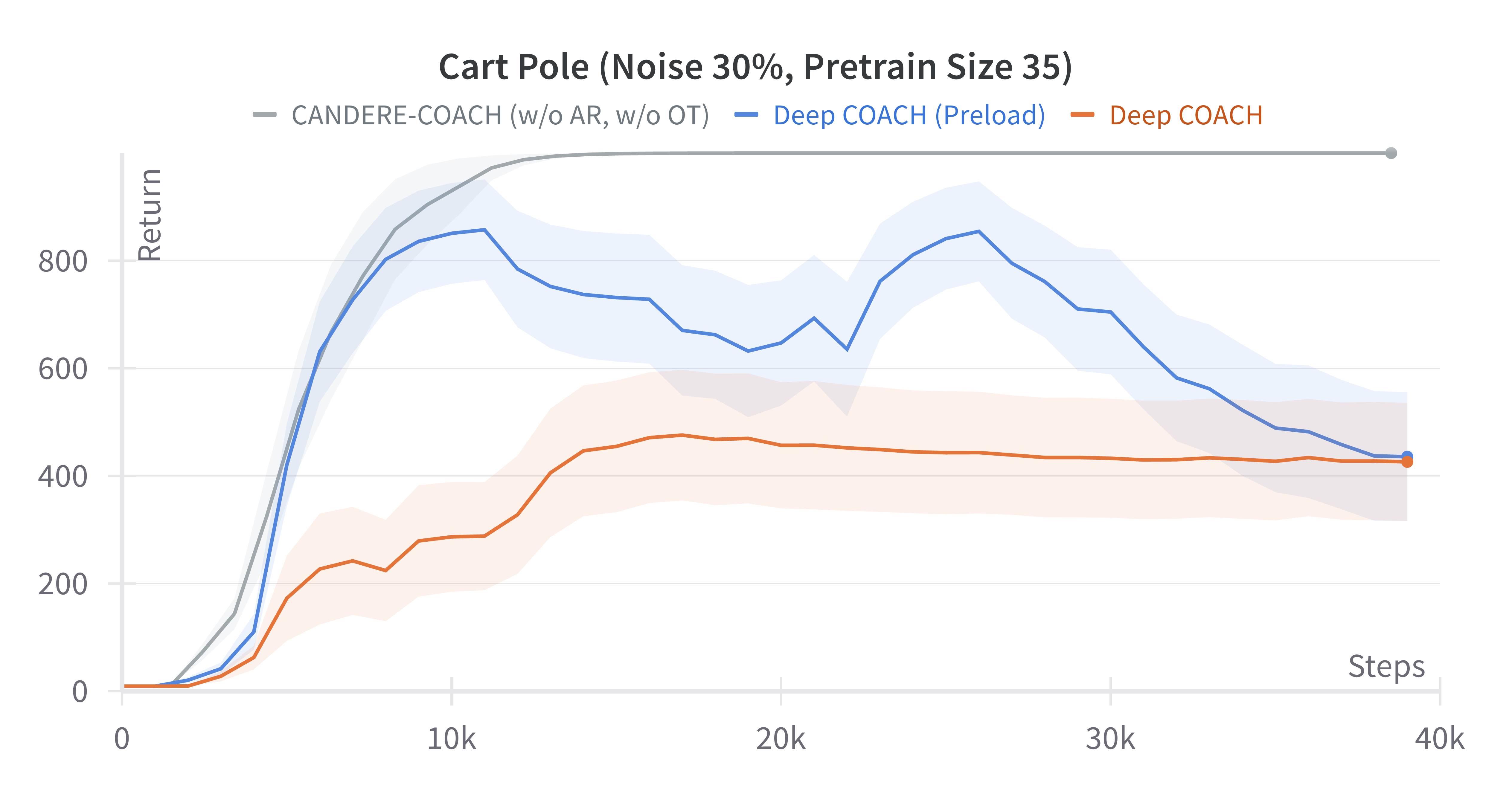}
\caption[Average return of CANDERE-COACH (w/o OT, w/o AR) in Cart Pole compared with Deep COACH in 30\% noise]{Ablation study on pertaining sizes: average episode return of \alg and Deep COACH in Cart Pole under 30\% noise. Three figures show the same experiment with different amounts of pretraining feedback. It can be seen that \alg needs at least 30 to succeed and 35 to perform well.}
\label{fig:reward_CP_CANOR_N30}
\end{figure}

\begin{figure}
  \centering
  \includegraphics[width=0.45\textwidth]{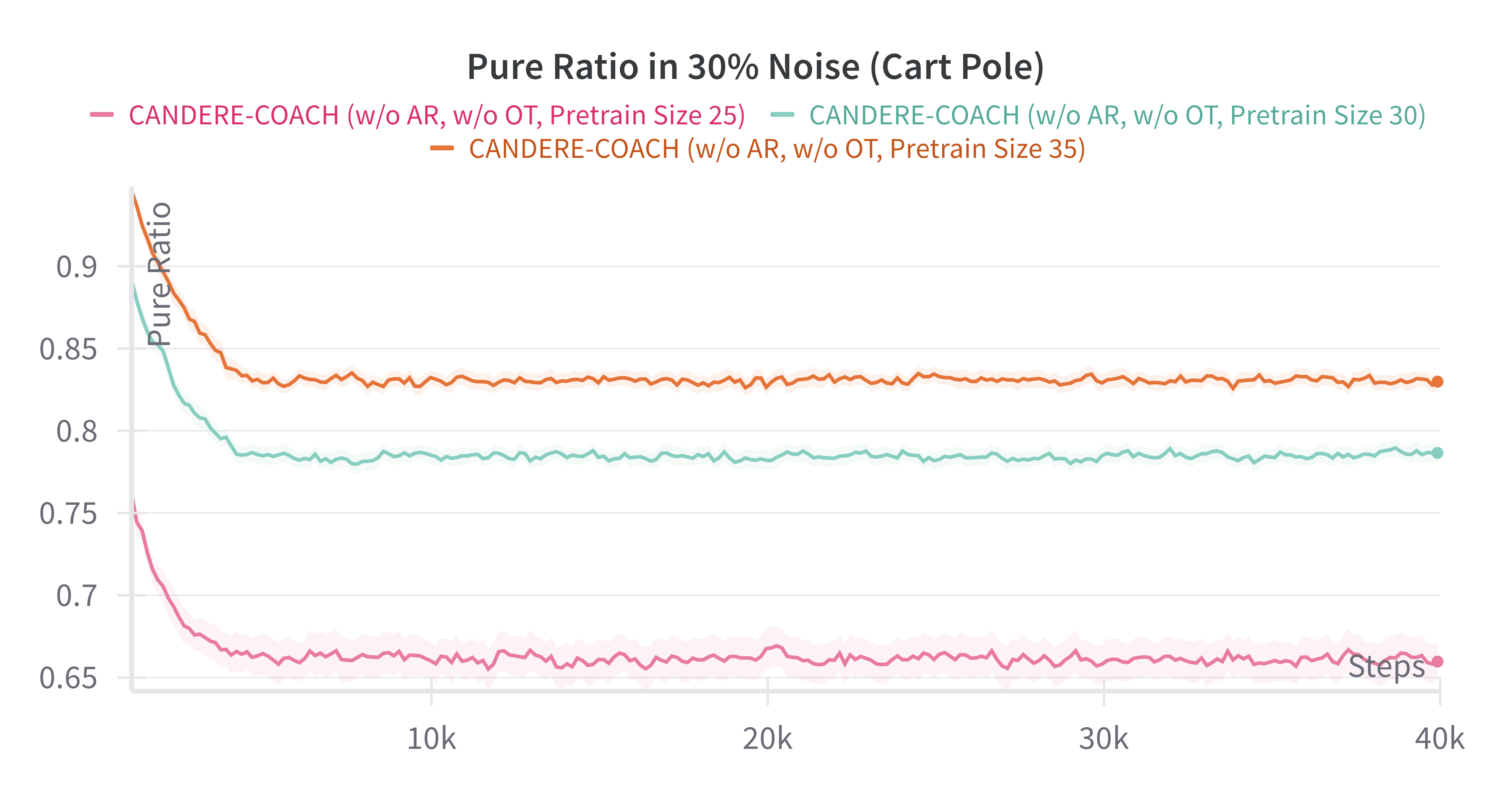}
\caption[Pure ratio of CANDERE COACH (w/o OT, w/o AR) in Cart Pole compared with Deep COACH in 30\% noise]{Ablation study on pertaining sizes: average pure ratio of \alg in CartPole under 30\% noise. While the agent explores new states and actions, the distribution of state and action changes and therefore a fixed classifier predicts less accurately over time.}
\label{fig:pure_ratio_CP_CANOR_N30}
\end{figure}

\begin{figure}
  \centering
  \includegraphics[width=0.45\textwidth]{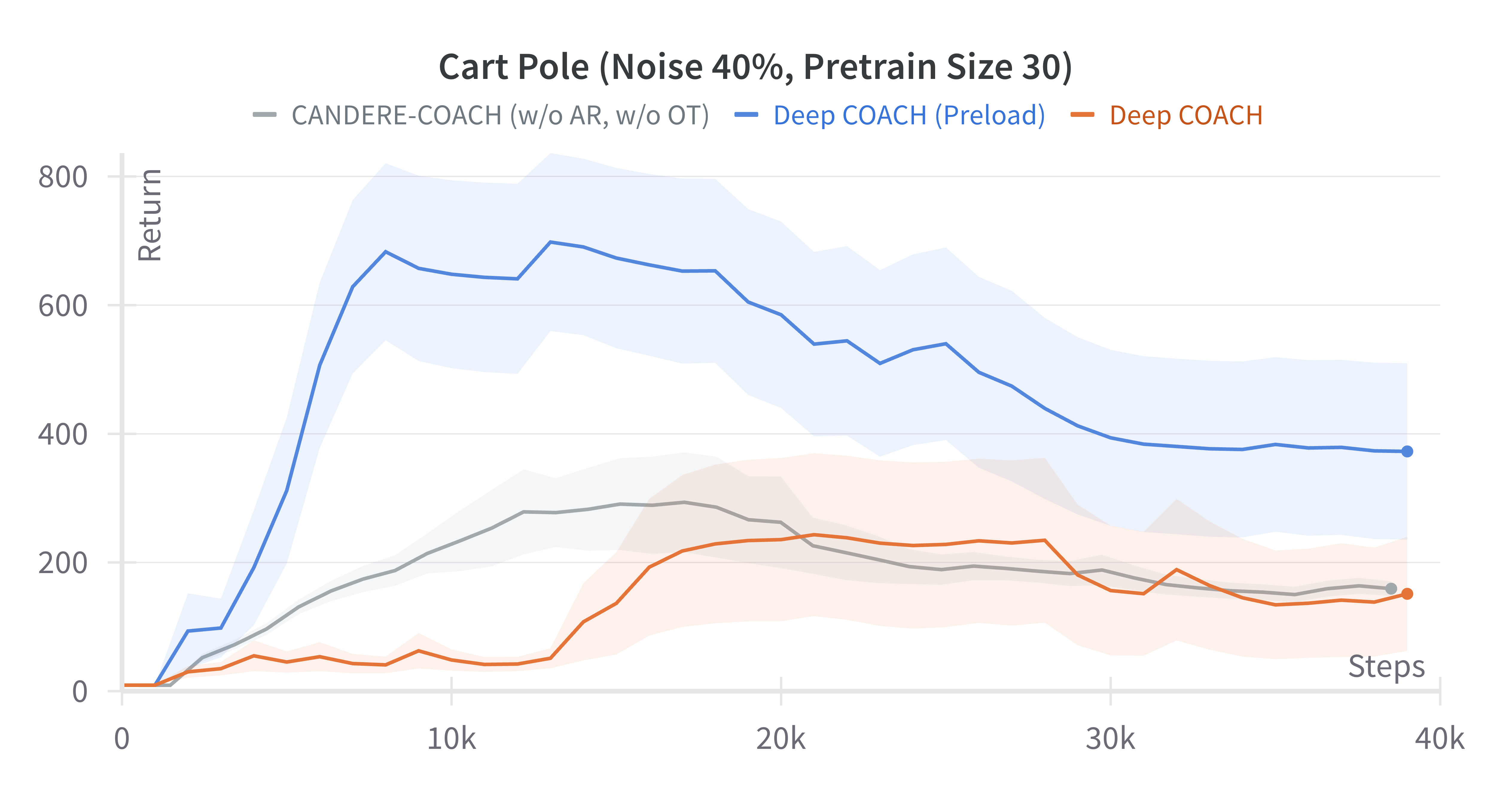}
  \includegraphics[width=0.45\textwidth]{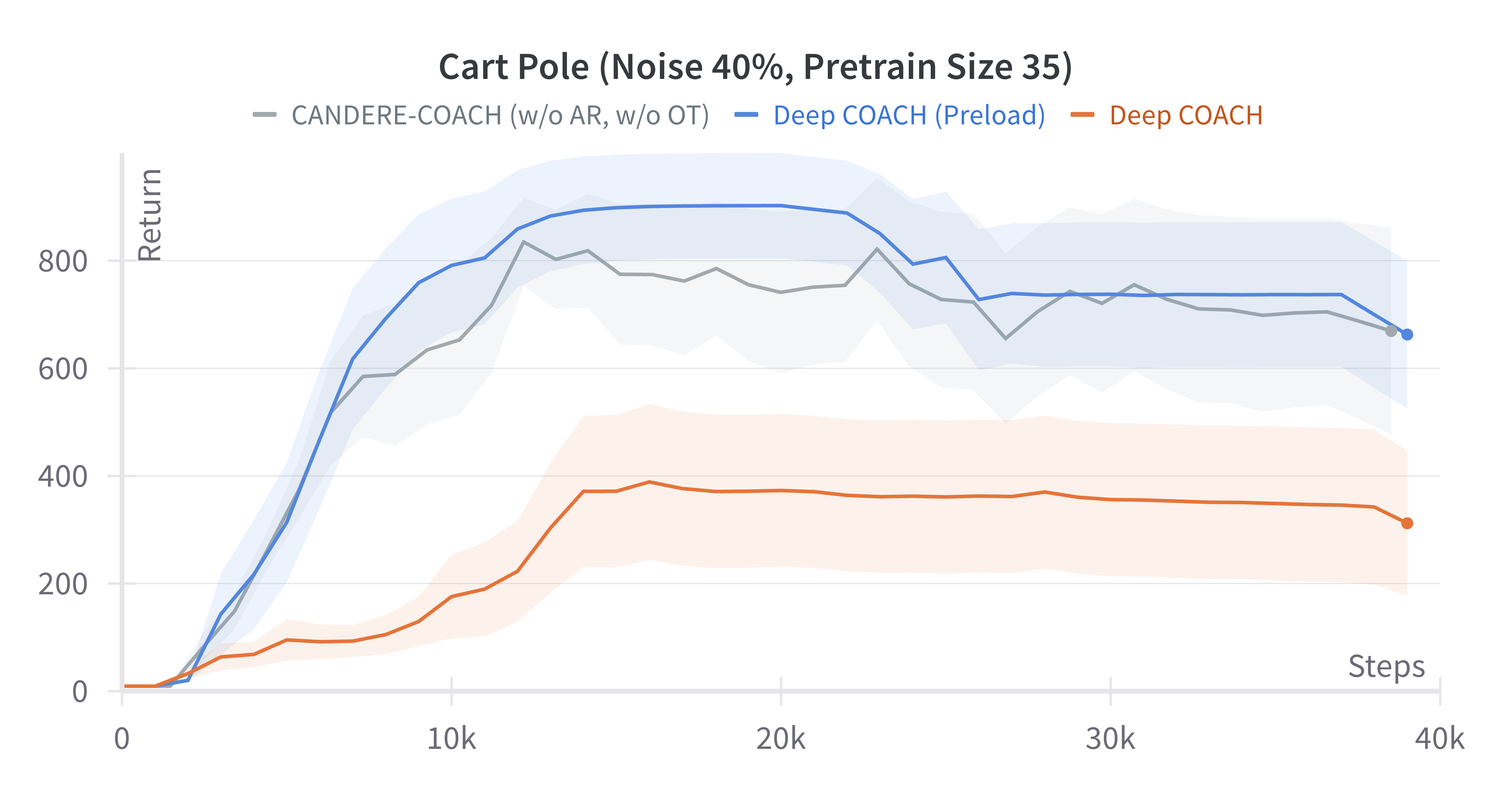}
    \includegraphics[width=0.45\textwidth]{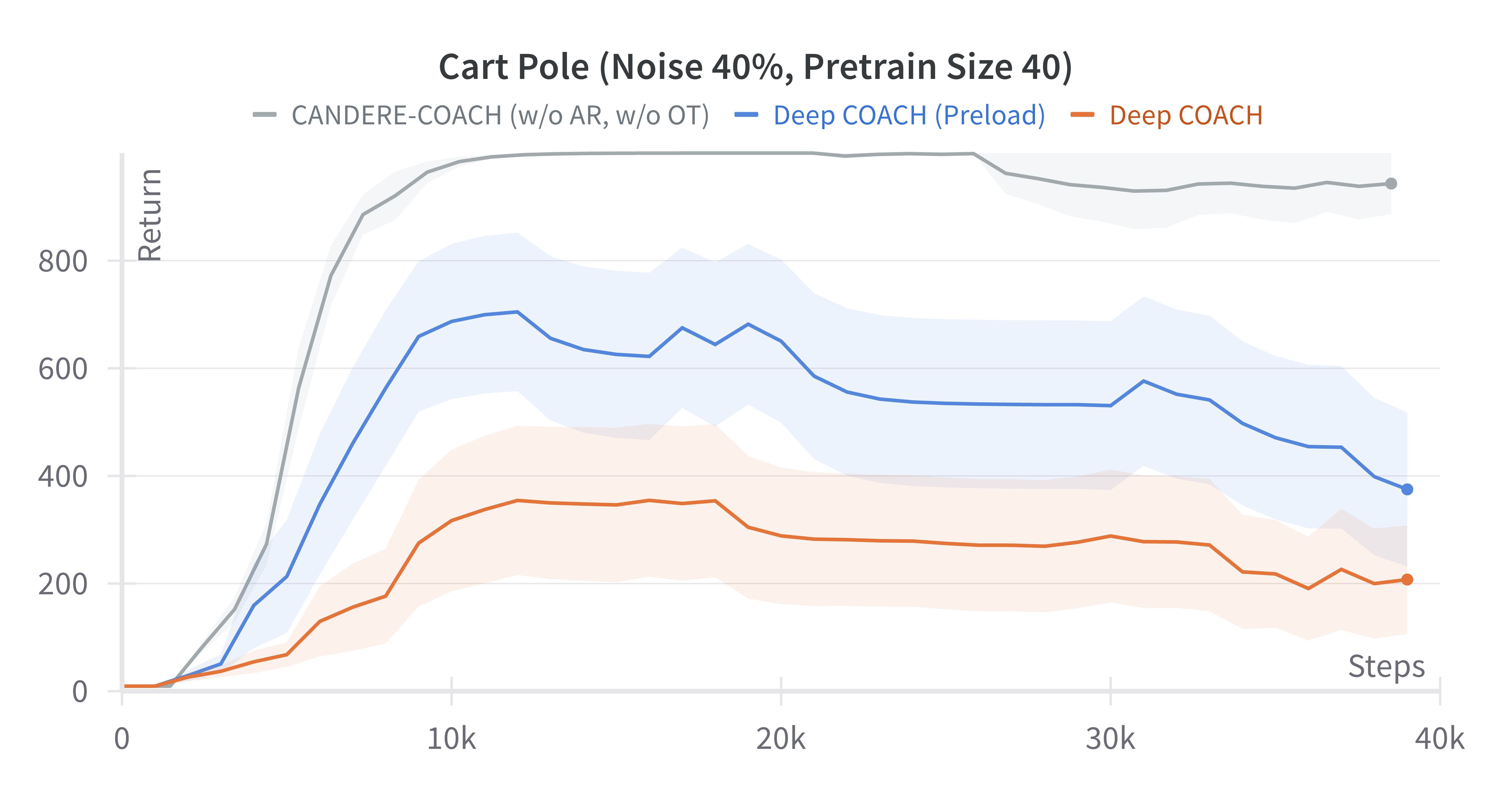}

\caption[Average return of CANDERE COACH (w/o OT, w/o AR) in Cart Pole compared with Deep COACH in 40\% noise]{Ablation study on pertaining sizes: average episode return of \alg and Deep COACH in Cart Pole under 40\% noise. Three figures show the same experiment with different amounts of pretraining feedback. 40\% is much harder and our \alg will also suffer from noise with pretraining size of 30 and \alg needs 40 to reach the maximum episodic return (1000).}
\label{fig:reward_CP_CANOR_N40}
\end{figure}

As shown in Figure~\ref{fig:reward_CP_CANOR_N30}, our method can learn well against 30\% noise, while Deep COACH shows highly unstable performance.
In fact, even with different amounts of preloaded pure dataset, under the same settings, Deep COACH shows a significantly different learning curve due to the very unpredictability of the noise. 
Furthermore, it is revealed in Figure~\ref{fig:pure_ratio_CP_CANOR_N30} that a classifier pretrained with datset of size 25 barely filters noise successfully as the pure ratio hovers around 70\% under 30\% noise, while the other two classifiers successfully improve the pure ratio up to 80\% and 85\%.
We also tested \alg on 40\% noise, as shown in Figure~\ref{fig:reward_CP_CANOR_N40}. 
With extremely high noises, \alg requires more pretraining data (35 in both 40\%) to succeed.
We can also observe a similar pattern in pure ratio which keeps decreasing and then becomes stable.

\begin{figure}[htbp!]
  \centering
  \includegraphics[width=0.45\textwidth]{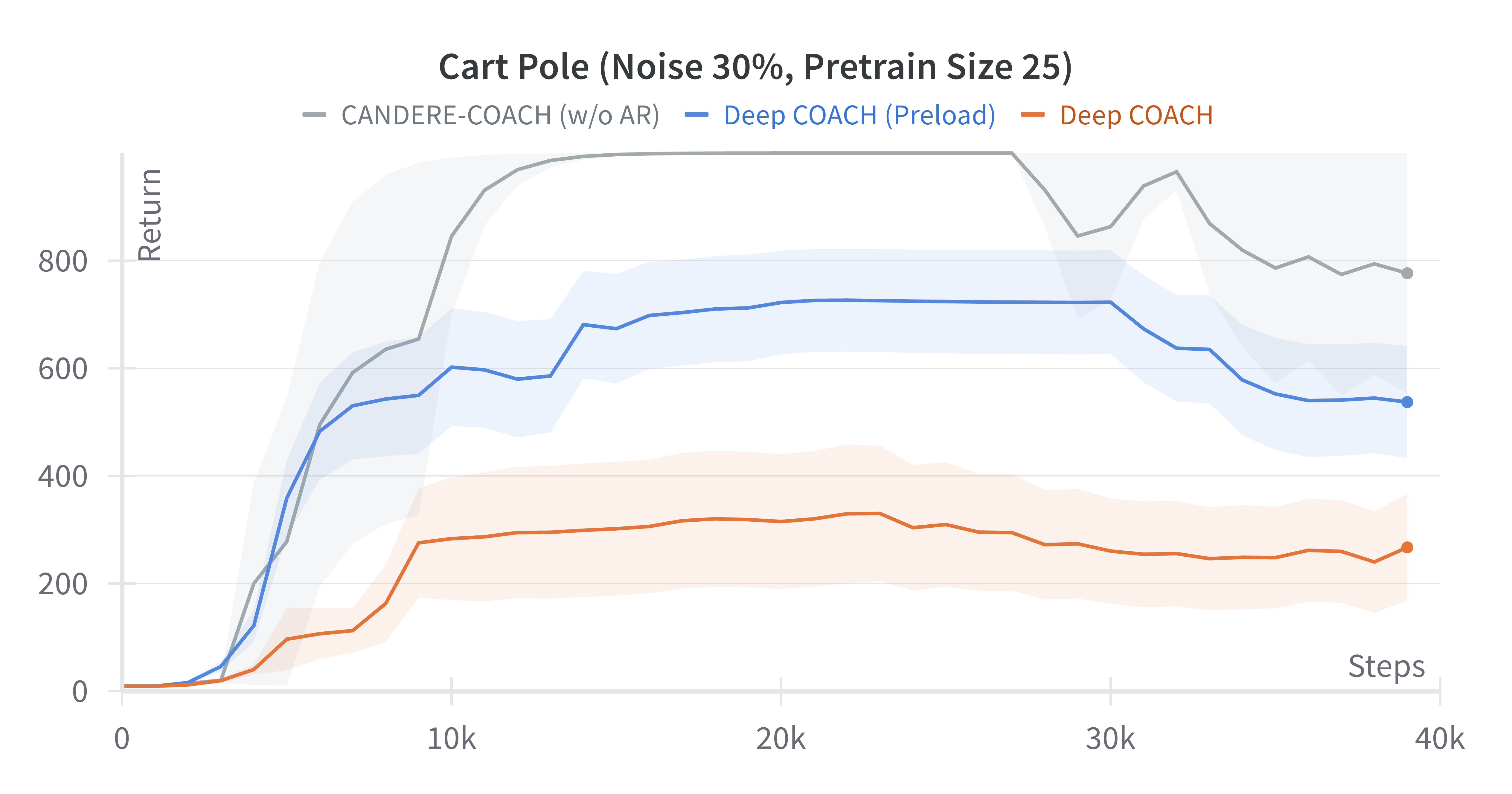}
  \includegraphics[width=0.45\textwidth]{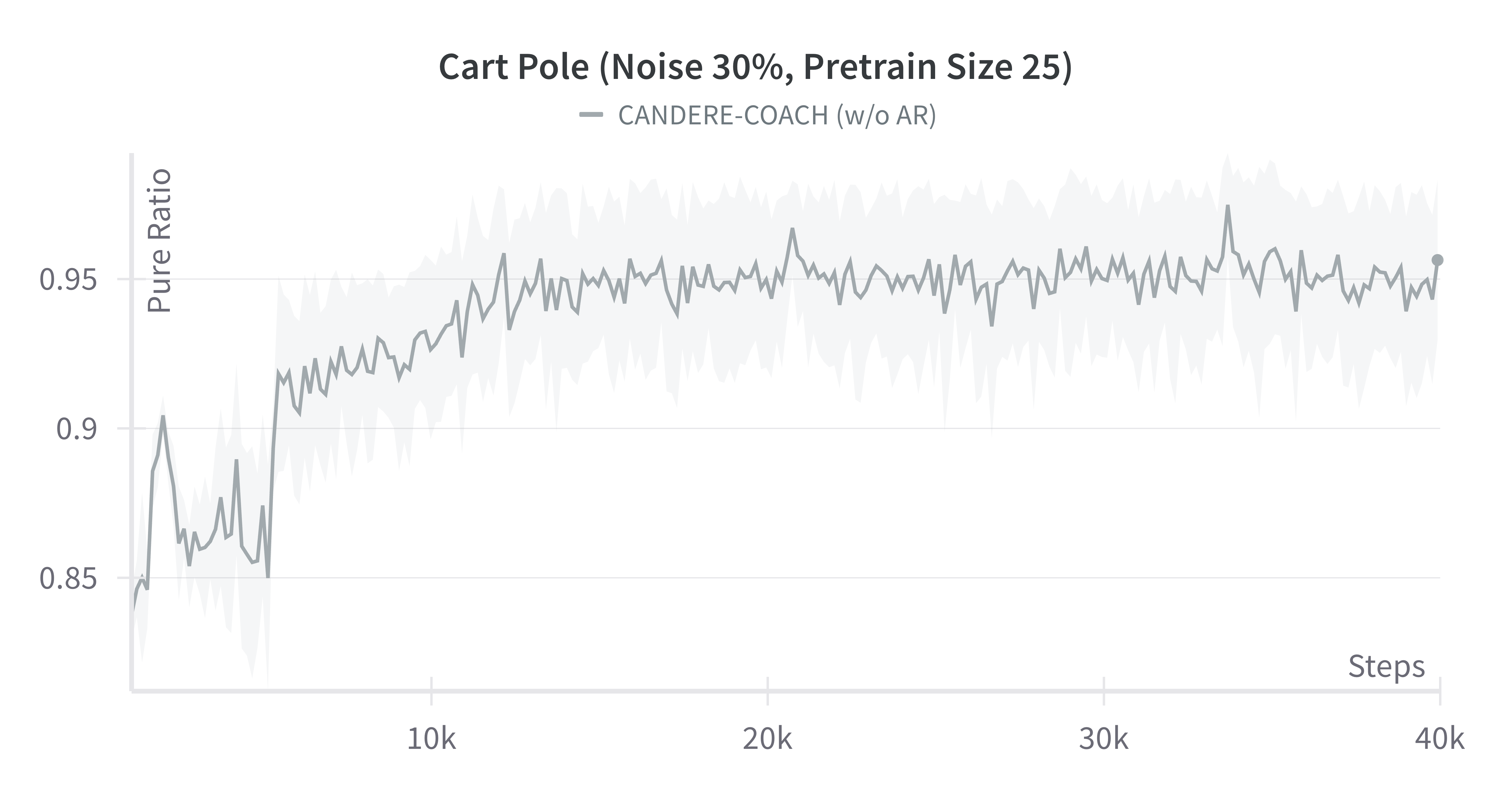}
\caption[Average return of CANDERE-COACH (w/o AR) in Cart Pole compared with Deep COACH in 30\% noise]{Performance of \alg with online training in 30\% noise. With online training, \alg is able to learn against 30\% noise with merely a pretraining dataset of size 25. Furthermore, its pure ratio successfully increases over time and reaches 95\%, while \alg without online training decreases over time.}
\label{fig:canor_ot_cp_n30}
\end{figure}

\begin{figure}[htbp!]
  \centering
  \includegraphics[width=0.45\textwidth]{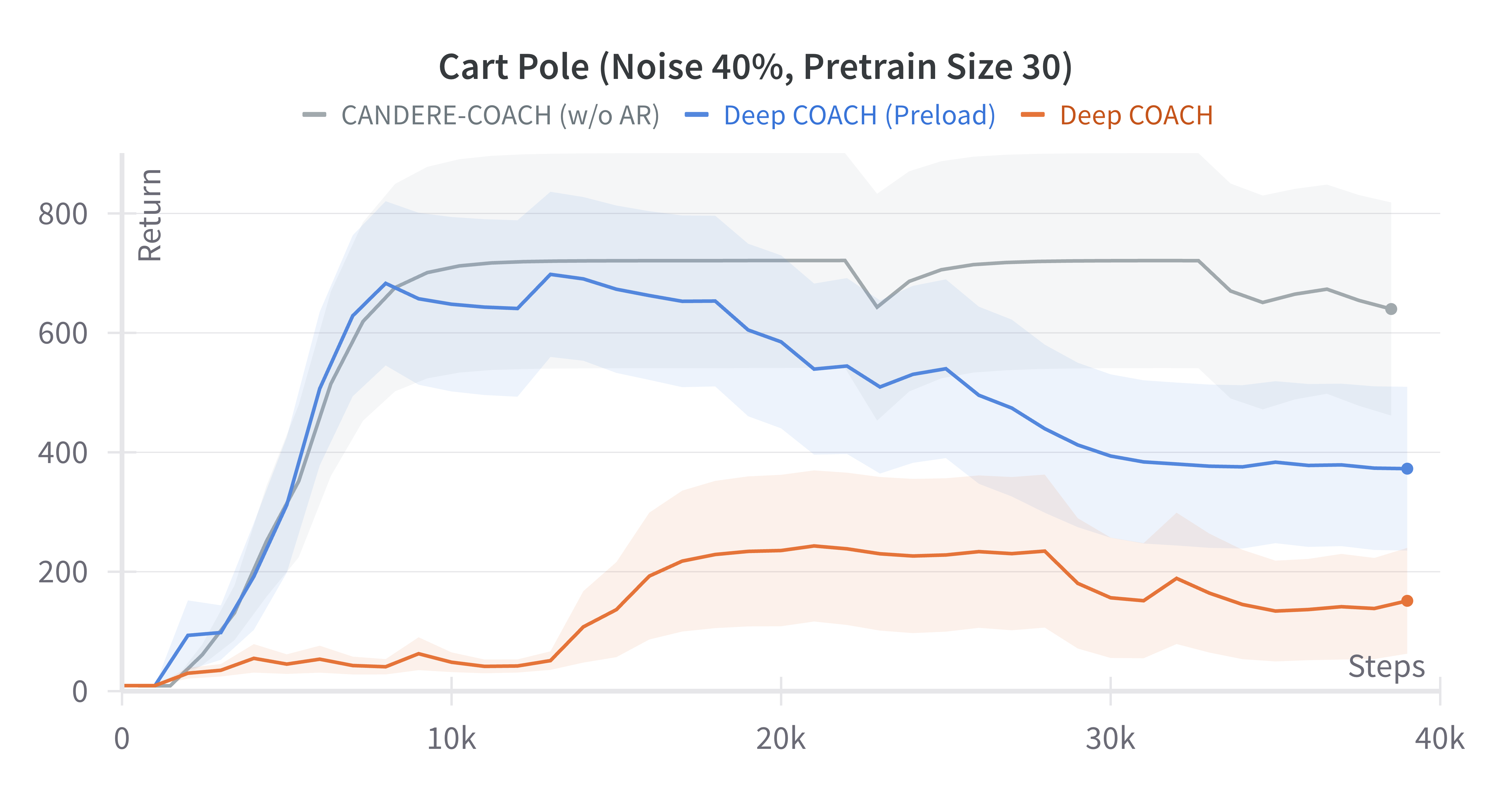}
    \includegraphics[width=0.45\textwidth]{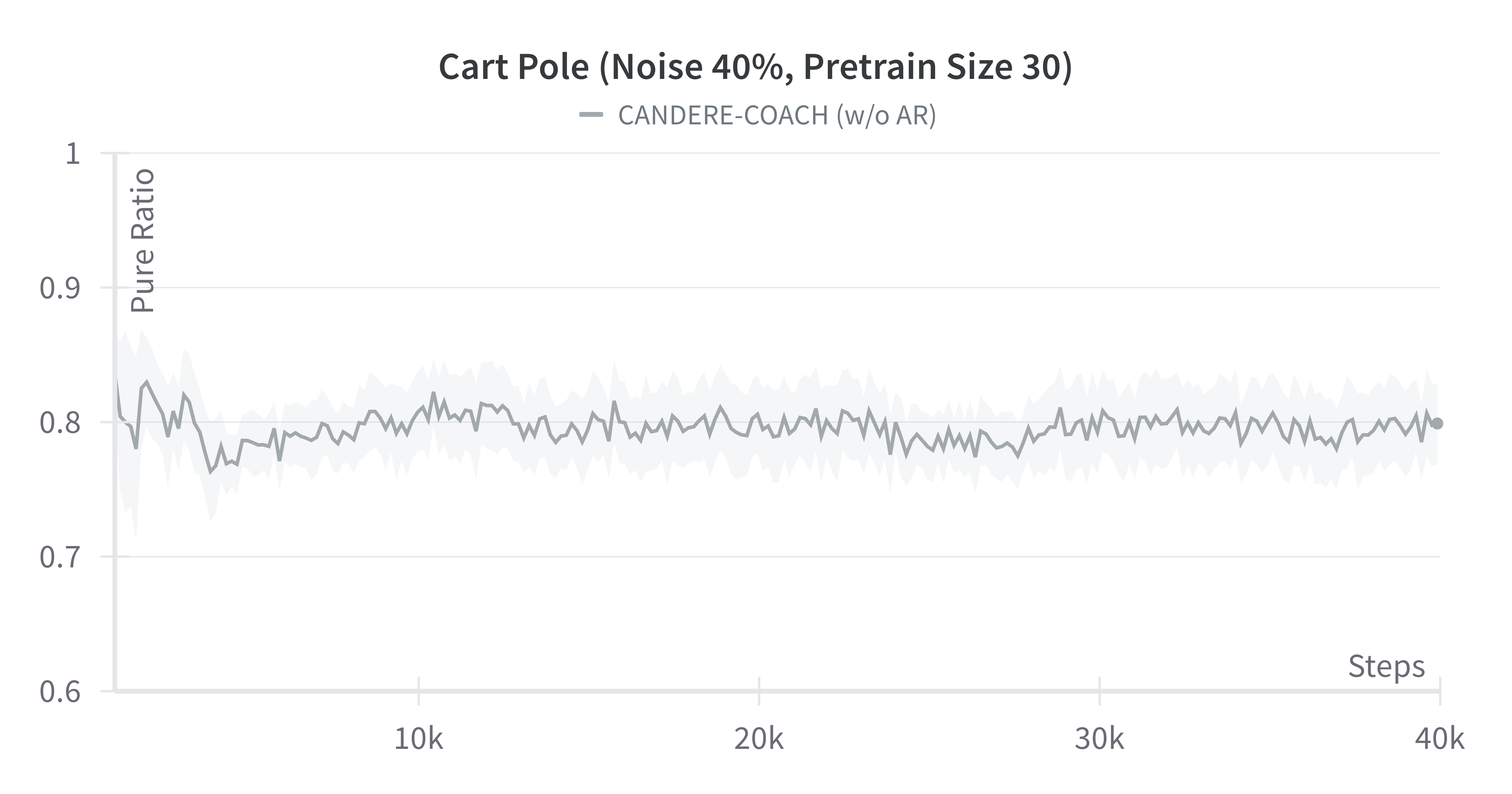}

\caption[Average return of CANDERE-COACH (w/o AR) in Cart Pole compared with Deep COACH in 40\% noise]{Performance of \alg with online training in 40\% noise.}
\label{fig:canor_ot_cp_n40}
\end{figure}

The results of CANDERE-COACH (w/o AR) in 30\% noise can be found in Figure~\ref{fig:canor_ot_cp_n30}.
Recall that in the previous section, \alg fails to learn with a pretraining dataset of size 25.
With online training, we observe that the pure ratio can gradually increase, which means that the classifier is learning and is well adapted to the new state and action distribution.
Eventually, the pure ratio stabilises around 95\% and as a result, \alg with online training is able to learn robustly against 30\% noise with pretraining dataset of size 25.
However, online training does not always show such a great improvement when noise increases.
The results in 40\% noise can be found in Figure~\ref{fig:canor_ot_cp_n40}.
Under 40\% noise, \alg with online training fails with 25 pretraining data and needs 30 to succeed. 
We also observe that the pure ratio under extremely high noise will go down and then fluctuate, which differs from the pattern in Figure~\ref{fig:canor_ot_cp_n30} and  Figure~\ref{fig:pure_ratio_CP_CANOR_N30}, which shows that the online training mechanism tends to perform worse under high noise.

\section{CANDERE-COACH in other noise levels}
\label{sec:candere_coach_other_noise}
\begin{figure*}[h]
  \centering
  \subfigure[Cart Pole]
  {\label{fig:n20_cp}
  \includegraphics[width=0.32\textwidth]{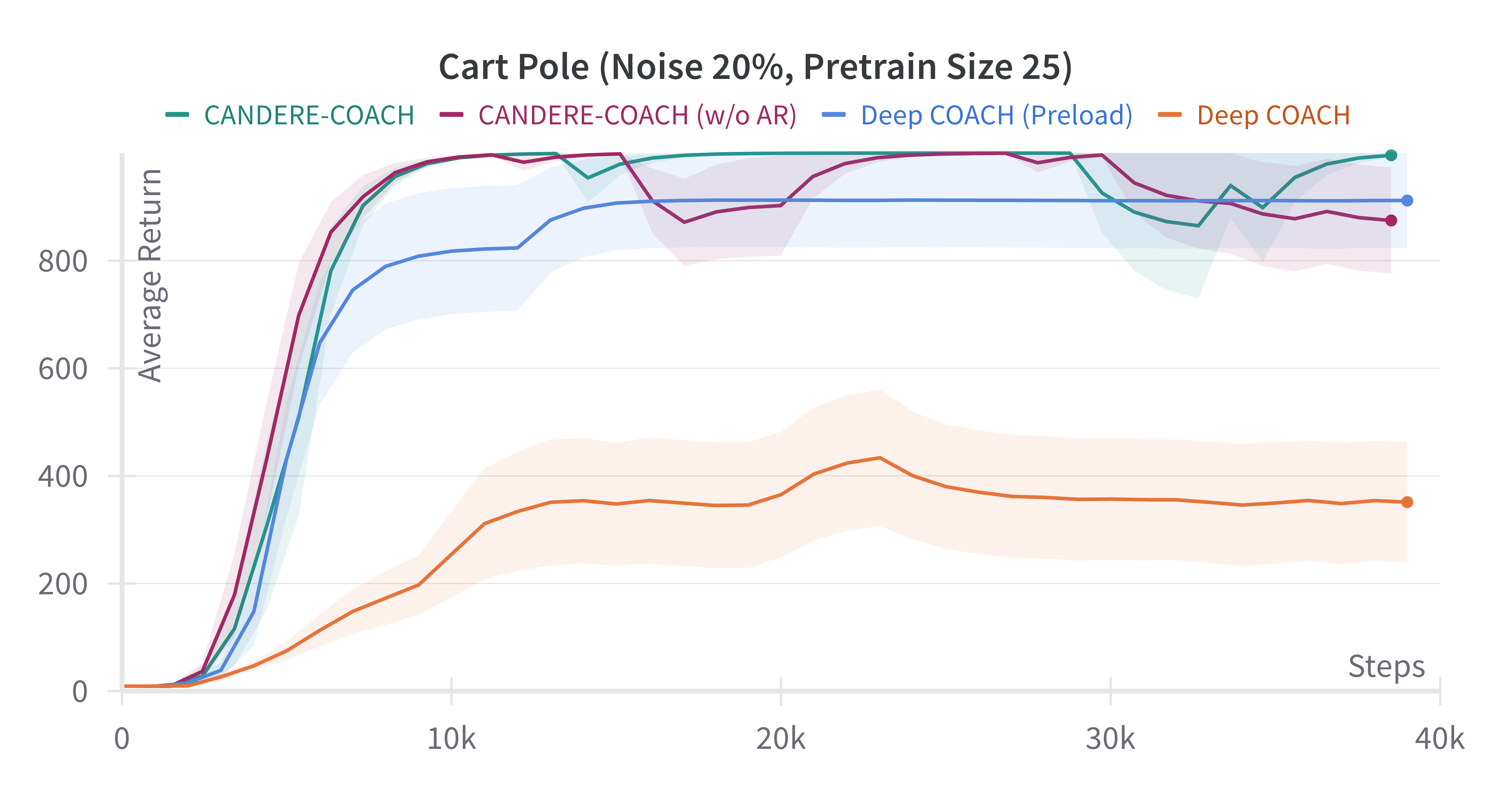}}
  \subfigure[Door key]
  {\label{fig:n20_dk}
  \includegraphics[width=0.32\textwidth]{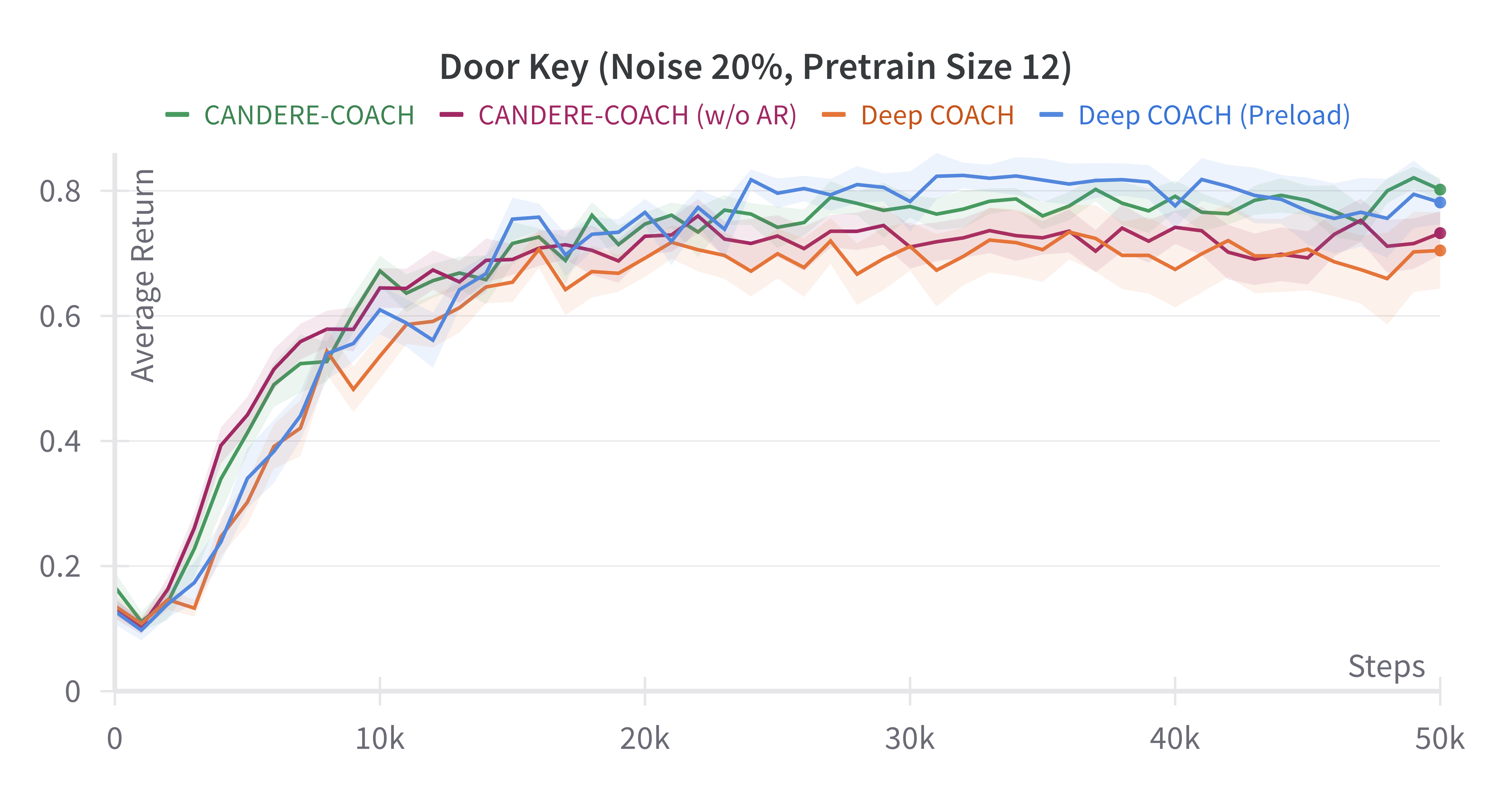}}
  \subfigure[Lunar Lander]
  {\label{fig:n20_ll}
  \includegraphics[width=0.32\textwidth]{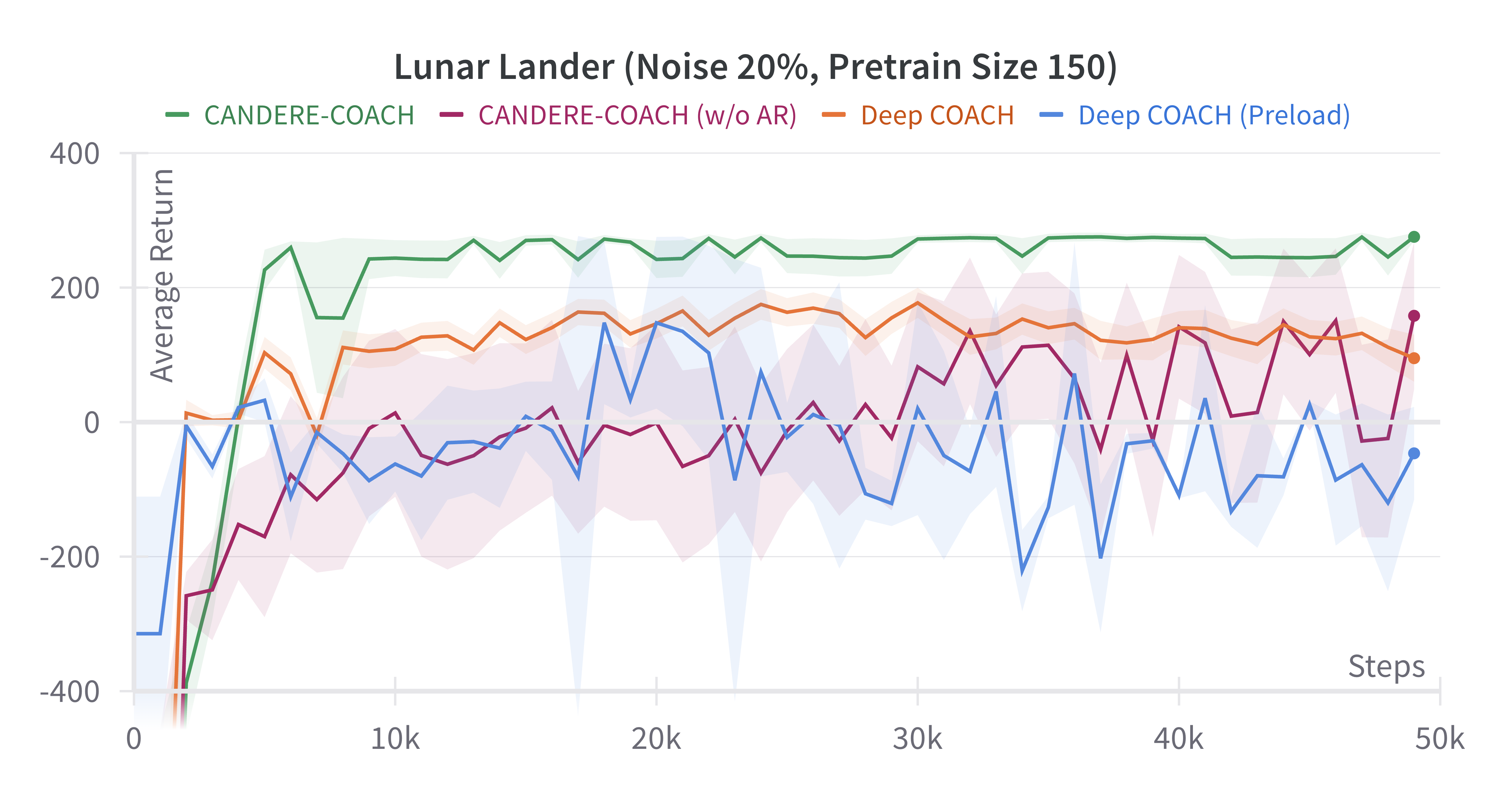}}
\caption{Performance comparison in 20\% noise in Cart Pole, Door Key and Lunar Lander}
\label{fig:n20}
\end{figure*}

We also conduct experiments in lower noise level like 20\%, the results are shown in Figure~\ref{fig:n20}.
With lower noise, in Cart Pole and Door Key, the performance difference is less significant because our baseline Deep COACH (Preload) can also perform well.
However, Lunar Lander is our hardest domain, and we still observe a similar pattern in 20\% noise, where CANDERE-COACH shows best performance and reaches 200 in episodic average return, while CANDERE-COACH (w/o AR) and baselines fail to do so.

\section{Study on feedback budget}

We conduct most of our experiments with a limited budget.
As different budgets can lead to different results, the budget for experiments is chosen based on such criteria: the budget should at least allow the agent to solve the task with 0\% noise, while the agent may fail with higher noise levels.

\begin{figure*}[h]
  \centering
  \subfigure[Noise 0\%]
  {\label{fig:budget_0}
  \includegraphics[width=0.32\textwidth]{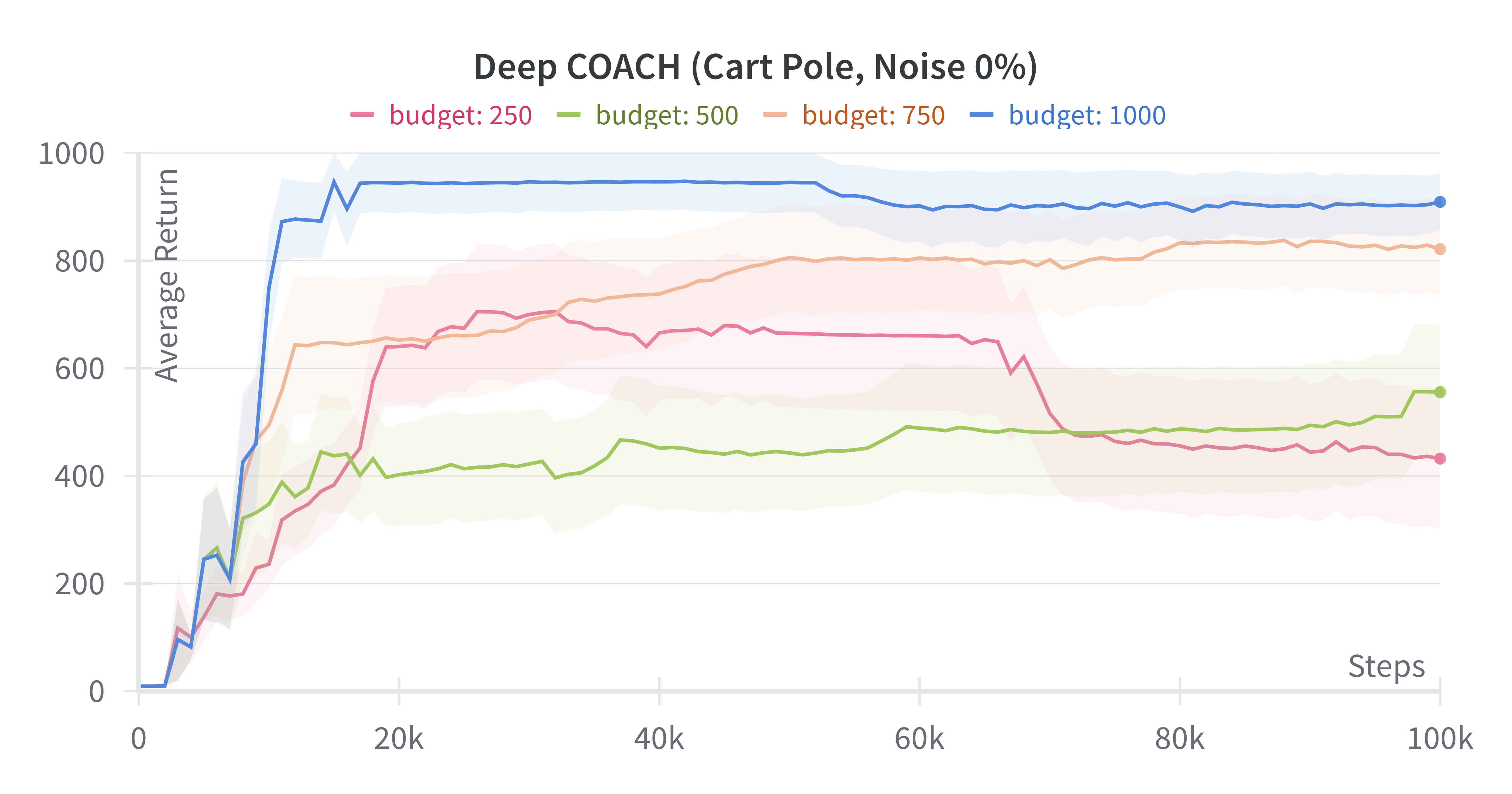}}
  \subfigure[Noise 10\%]
  {\label{fig:budget_10}
  \includegraphics[width=0.32\textwidth]{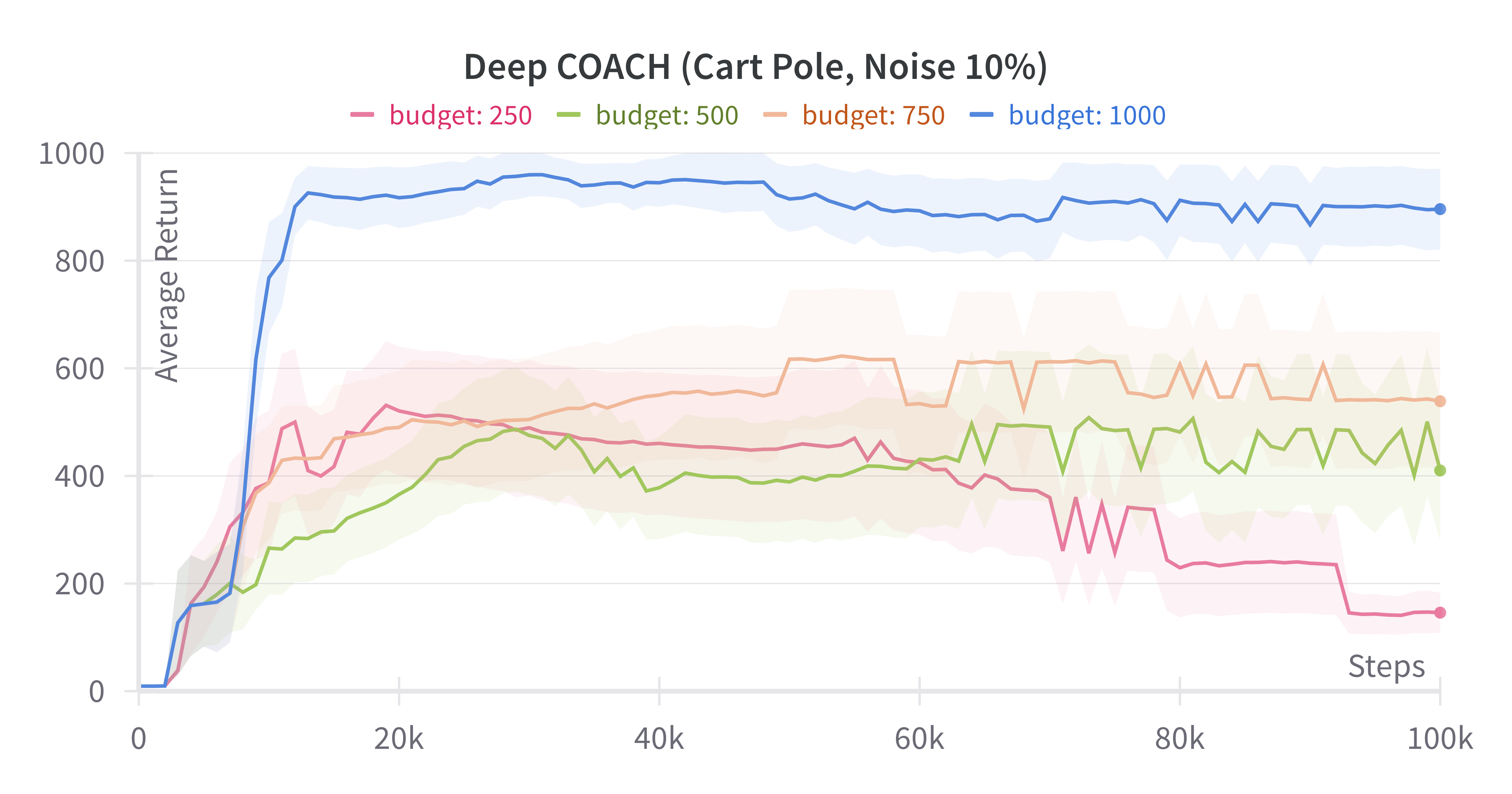}}
  \subfigure[Noise 20\%]
  {\label{fig:budget_20}
  \includegraphics[width=0.32\textwidth]{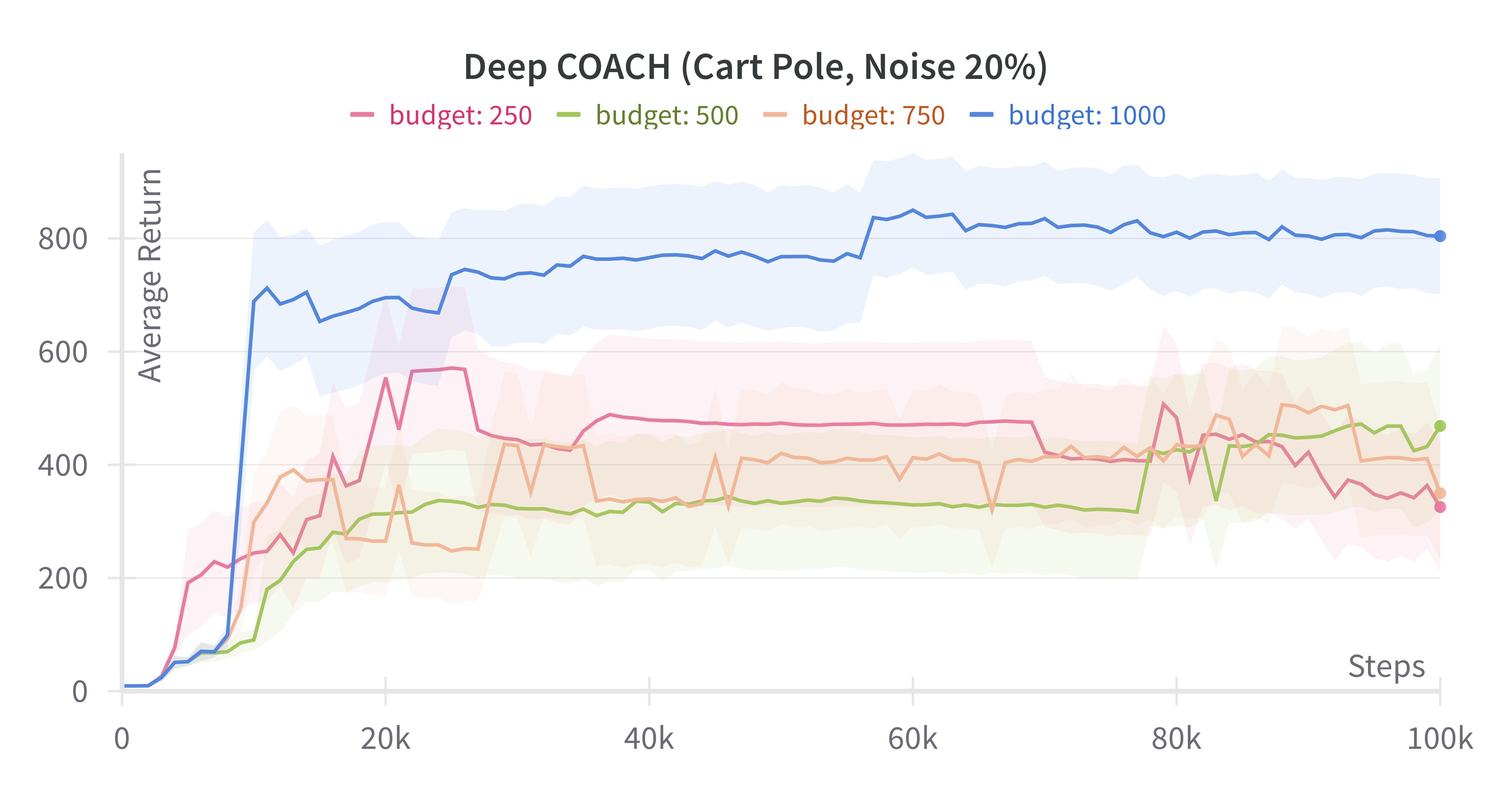}}
  \subfigure[Noise 30\%]
  {\label{fig:budget_30}
  \includegraphics[width=0.32\textwidth]{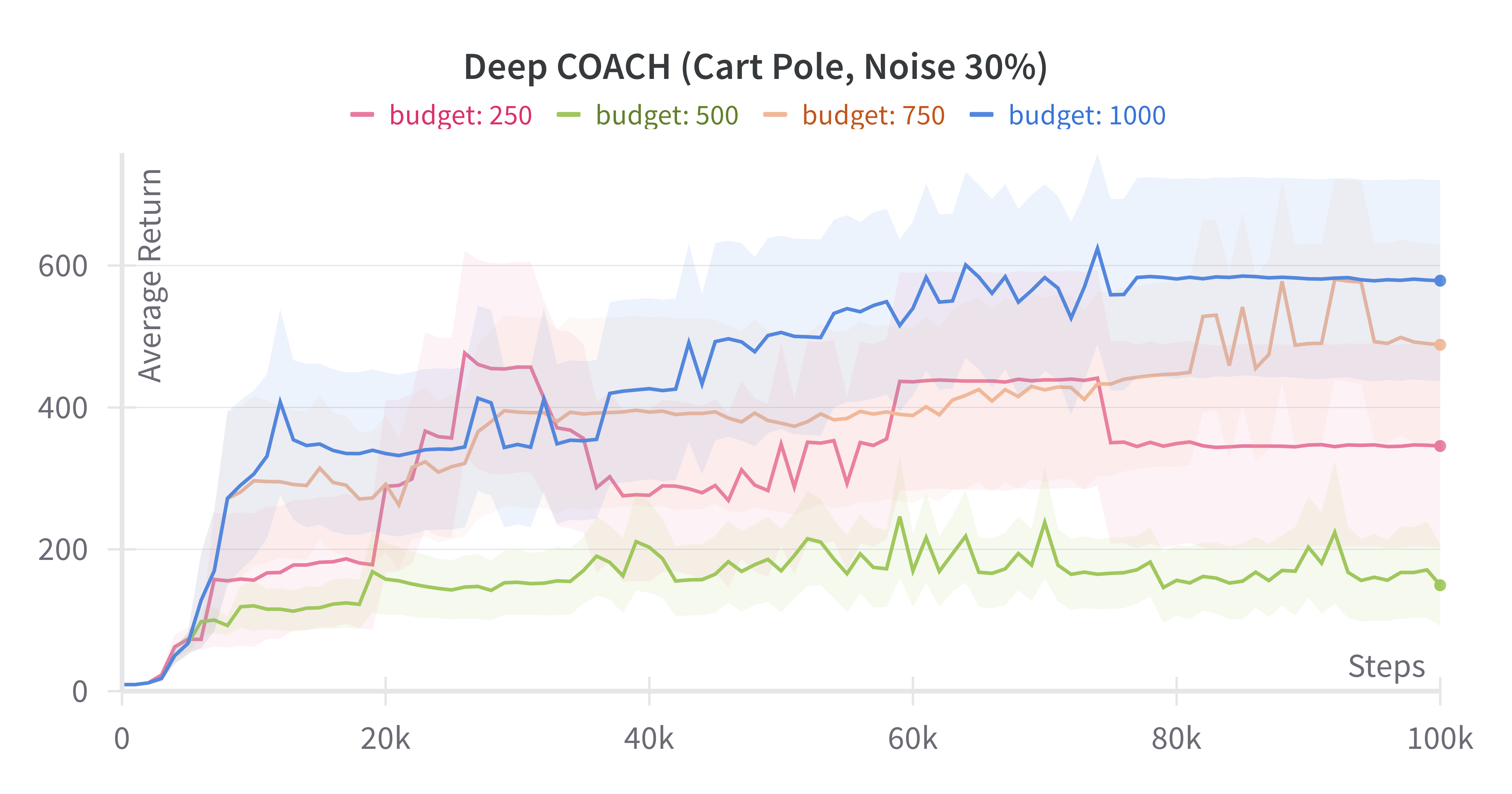}}
  \subfigure[Noise 40\%]
  {\label{fig:budget_40}
  \includegraphics[width=0.32\textwidth]{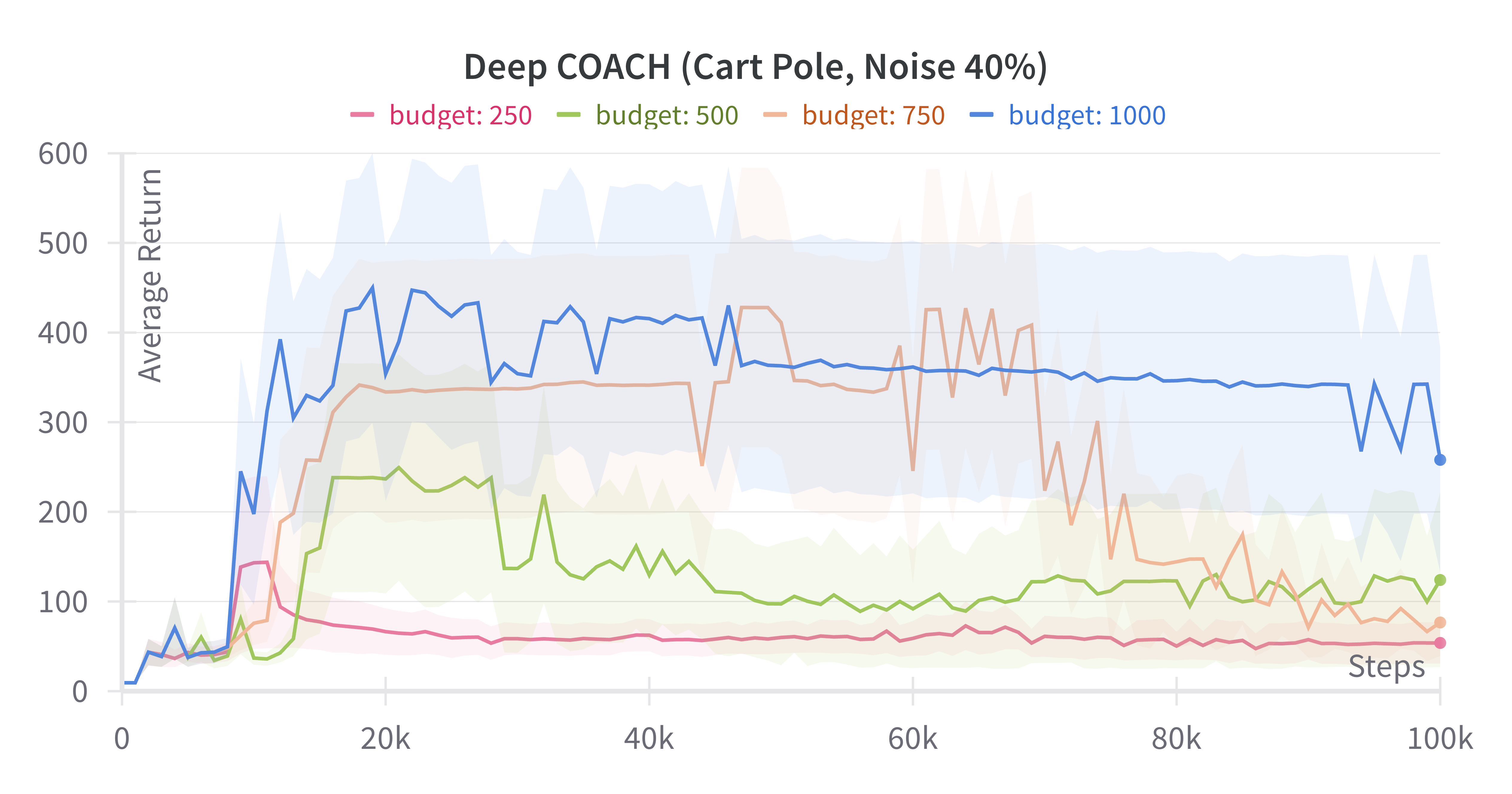}}
  
\caption{Performance of Deep COACH (Preload) under different scales of noise and budget in Cart Pole }
\label{fig:budget_study}
\end{figure*}

A study on budget's influence on Deep COACH with a noise-free budget of 30 is shown in Figure~\ref{fig:budget_study}. 
A budget of 1000 allows the agent to reach very close to maximum episodic return smoothly, while the performance drops as noise increases.
With a smaller budget like 750, the agent fails to do so and hence we choose 1000 as the budget for Cart Pole in our experiments for a comparison between Deep COACH and CANDERE-COACH. 

\section{Scripted teacher}
\label{sec:scripted_teacher}
Our scripted teacher is trained with PPO \cite{schulman2017proximal} following hyperparameters of RL Zoo \cite{rl-zoo3}. More details are described in the following subsections for each domain.
\subsection{Cart Pole} 
In Cart Pole, the expert is trained with PPO with hyperparameters shown in Table~\ref{tab:cp_ppo_hyperparam}.
\begin{table}[!htbp]
    \centering
    \begin{tabular}{cc}
        \hline\hline
        Hyperparameter & Value\\
        \hline\hline
        Time steps & 1e5\\
        Rollout steps & 32\\
        Gae lambda & 0.8\\
        Gamma & 0.9\\
        Learning rate & 1e-3 \\
        Clip range & 0.2\\
        Batch size & 256\\
        Hidden units & 64 \\
        Layers & 2 \\
        Activation & ReLU \\
        \hline\hline

    \end{tabular}
    \caption{Hyperparamters of Cart Pole expert training}
    \label{tab:cp_ppo_hyperparam}
\end{table}

\subsection{Door Key}
In Door Key, the expert is trained with PPO with hyperparameters shown in Table~\ref{tab:dk_ppo_hyperparam}.
Furthermore, Minigrid Doorkey is set to be fully observable and we use a CNN-based feature extractor to reduce the observation space to 5.

\begin{table}[!htbp]
    \centering
    \begin{tabular}{cc}
        \hline\hline

        Hyperparameter & Value\\
        \hline\hline
        Time steps & 1e5\\
        Rollout steps & 128\\
        Gae lambda & 0.95\\
        Gamma & 0.99\\
        Learning rate & 2.5e-4 \\
        Clip range & 0.2\\
        Batch size & 64\\
        Hidden units & 64 \\
        Layers & 2 \\
        Activation function & ReLU \\
        CNN channels & [16, 32, 64] \\
        CNN kernel size & \(2, 2\) \\
        \hline\hline

    \end{tabular}
    \caption{Hyperparamters of Door Key expert training}
    \label{tab:dk_ppo_hyperparam}
\end{table}

\subsection{Lunar Lander}
In Lunar Lander, the expert is trained with PPO with hyperparameters shown in Table~\ref{tab:ll_ppo_hyperparam}.
\begin{table}[!htbp]
    \centering
    \begin{tabular}{cc}
        \hline\hline
        Hyperparameter & Value\\
        \hline\hline
        Time steps & 1e6\\
        Rollout steps & 1024\\
        Gae lambda & 0.98\\
        Gamma & 0.999\\
        Learning rate & 1e-3 \\
        Clip range & 0.2\\
        Batch size & 64\\
        Hidden units & 64 \\
        Layers & 2 \\
        Activation & ReLU \\
        \hline\hline

    \end{tabular}
    \caption{Hyperparamters of Lunar Lander expert training}
    \label{tab:ll_ppo_hyperparam}
\end{table}

\section{Collecting the pretraining dataset}
The pretraining dataset is collected with our scripted teacher to label with state-action pairs positive or negative feedback. 
We first collect states following a certain distribution to ensure better coverage of the state space. 
Then we label the states and optimal actions to be positive and the collected dataset will be randomly shuffled.
Lastly, we practice data augmentation by labelling all the nonoptimal actions to be negative. 
The details of the state sampling distribution of each domain are described in the following subsections.

\subsection{Cart Pole}
In Cart Pole, the state is sampled uniformly in a clipped observation space.
The state space of Cart Pole consists of four dimensions and two of them are not bounded. Therefore, we properly choose a suitable clip range to sample the states. 
We set the sampling range of position and velocity based on the fact that an episode will be terminated if the cart leaves the $(-2.4, 2.4)$ range.
Details can be seen in Table~\ref{tab:cp_clip_obs}. 

\begin{table*}[!htbp]
    \centering
    \begin{tabular}{ccccc}
        \hline\hline
        Dimension & Original Min & Clipped Min & Original Max & Clipped Max\\
        \hline\hline
        Cart Position & -4.8 & -2.4 & 4.8 & 2.4\\
        Cart Velocity & -Inf & -2.4 & +Inf & 2.4\\
        Pole Angle & -0.418 & -0.418 & 0.418 & 0.418\\
        Pole Angular Velocity & -Inf & -0.418 & +Inf & 0.418\\
        \hline\hline

    \end{tabular}
    \caption{Sampling space of Cart Pole}
    \label{tab:cp_clip_obs}
\end{table*}

\subsection{Door Key \& Lunar Lander}
In Door Key and Lunar Lander, the dataset is sampled from trajectories following a sampling policy that takes expert action by 50\% chance and random action by 50\% chance.
The reason for this is different for these domains.
In Door Key, we cannot practice uniform sampling in the RGB image array space.
In Lunar Lander, the observation space is significantly larger and if we practice uniform sampling, most sampled states will never be visited by the agent and therefore bring low performance of the classifier.

\section{COACH Hyperparameters}
\label{sec:hyperparam}
In this section, we show the hyperparameters used in our experiments for CANDERE-COACH and Deep COACH. Deep COACH shares the same hyperparameters with CANDERE-COACH if applicable.
\subsection{Cart Pole}
The hyperparameters in Cart Pole can be seen in Table~\ref{tab:CANDERE_coach_hyperparam_cp}.
\begin{table}[!htbp]
    \centering
    \begin{tabular}{cc}
        \hline\hline
        Hyperparameter & Value\\
        \hline\hline
        Actor learning rate & 0.00005\\
        Batch size & 256\\
        Budget & 1000\\
        Eligibility trace windows size & 10 \\
        Eligibility trace decay factor & 0.35 \\
        Classifier learning rate & 0.01\\
        Feedback frequency & 10\\
        Actor hidden units & 1024 \\
        Actor layers & 2 \\
        Actor activation & ReLU \\
        Q function hidden units & 1024 \\
        Q function layers & 2 \\
        Q function activation & ReLU \\
        Classifier hidden units & 64 \\
        Classifier layers & 2 \\
        Classifier activation & ReLU \\
        Classifier pretraining epochs & 100 \\
        Classifier pretraining learning rate & 0.001 \\
        Classifier pretraining loss & Cross entropy loss \\
        Active relabelling rate & 0.6 \\
        \hline\hline

    \end{tabular}
    \caption{Hyperparamters of CANDERE-COACH in Cart Pole }
    \label{tab:CANDERE_coach_hyperparam_cp}
\end{table}

\subsection{Door Key}
The hyperparameters in Door Key can be seen in Table~\ref{tab:CANDERE_coach_hyperparam_dk}.
\begin{table}[!htbp]
    \centering
    \begin{tabular}{cc}
        \hline\hline
        Hyperparameter & Value\\
        \hline\hline
        Actor learning rate & 0.00005\\
        Batch size & 256\\
        Budget & 500\\
        Eligibility trace windows size & 10 \\
        Eligibility trace decay factor & 0.35 \\
        Classifier learning rate & 0.001\\
        Feedback frequency & 10\\
        Actor hidden units & 1024 \\
        Actor layers & 2 \\
        Actor activation & ReLU \\
        Q function hidden units & 1024 \\
        Q function layers & 2 \\
        Q function activation & ReLU \\
        Classifier hidden units & 64 \\
        Classifier layers & 2 \\
        Classifier activation & ReLU \\
        Classifier pretraining epochs & 100 \\
        Classifier pretraining learning rate & 0.001 \\
        Classifier pretraining loss & Focal loss \\
        Active relabelling rate & 0.8 \\
        \hline\hline

    \end{tabular}
    \caption{Hyperparamters of CANDERE-COACH in Cart Pole }
    \label{tab:CANDERE_coach_hyperparam_dk}
\end{table}

\subsection{Lunar Lander}
The hyperparameters in Lunar Lander can be seen in Table~\ref{tab:CANDERE_coach_hyperparam_ll}.
\begin{table}[!htbp]
    \centering
    \begin{tabular}{cc}
        \hline\hline
        Hyperparameter & Value\\
        \hline\hline
        Actor learning rate & 0.00005\\
        Batch size & 256\\
        Budget & 5000\\
        Eligibility trace windows size & 10 \\
        Eligibility trace decay factor & 0.35 \\
        Classifier learning rate & 0.001\\
        Feedback frequency & 10\\
        Actor hidden units & 1024 \\
        Actor layers & 2 \\
        Actor activation & ReLU \\
        Q function hidden units & 1024 \\
        Q function layers & 2 \\
        Q function activation & ReLU \\
        Classifier hidden units & 64 \\
        Classifier layers & 2 \\
        Classifier activation & ReLU \\
        Classifier pretraining epochs & 100 \\
        Classifier pretraining learning rate & 0.001 \\
        Classifier pretraining loss & Focal loss \\
        Active relabelling rate & 0.6 \\
        \hline\hline

    \end{tabular}
    \caption{Hyperparamters of CANDERE-COACH in Lunar Lander}
    \label{tab:CANDERE_coach_hyperparam_ll}
\end{table}

\subsection{Summary on variations of CANDERE-COACH}
We summarise the aforementioned CANDERE-COACH and its variations' domain-wise performance and required amount of pretraining dataset size, as shown in Table~\ref{tab:summary}.

\begin{table*}[h]
    \centering
    \resizebox{\textwidth}{!}
    {
        \begin{tabular}{|c|c|c|c|}
        \hline
        \diagbox[]{Domain \&Noise}{Algorithm} &CANDERE-COACH (w/o AR, w/o OT) & CANDERE-COACH (w/o AR) & CANDERE-COACH\\
        \hline
        Cart Pole, 30\% & Outperform with pretrain size 30  & Outperform with pretrain size 25 & Outperform with pretrain size 20 \\
        \hline
        Cart Pole, 40\% & Outperform with pretrain size 30 & Outperform Outperform with pretrain size 30 & Outperform with pretrain size 25\\
        \hline
        Door Key, 30\% & Fail with pretrain size 12  & Outperform with pretrain size 12 & Outperform with pretrain size 12 \\
        \hline
        Door Key, 40\% & Fail with pretrain size 12 & Fail with pretrain size 12 & Outperform with pretrain size 12\\
        \hline
        Lunar Lander, 30\% & Fail with pretrain size 150  & Fail with pretrain size 150 & Outperform with pretrain size 150 \\
        \hline
        Lunar Lander, 40\% & Fail with pretrain size 150  & Fail with pretrain size 150 & Outperform with pretrain size 150 \\
        \hline
        \end{tabular}
    }
    \caption{Summary of results of CANDERE-COACH and its variations}
    \label{tab:summary}
\end{table*}

\end{document}